\documentclass[10pt,journal,compsoc]{IEEEtran}
\ifCLASSOPTIONcompsoc
\usepackage[nocompress]{cite}
\else
\usepackage{cite}
\fi

\ifCLASSINFOpdf
\usepackage[pdftex]{graphicx}
\usepackage{multirow}
\usepackage{amsmath}
\usepackage{amssymb}
\usepackage{caption}
\newcommand{\tabincell}[2]{\begin{tabular}{@{}#1@{}}#2\end{tabular}}
\usepackage{xcolor} 
\usepackage{color} 
\usepackage{verbatim} 
\else
\fi

\begin{document}
	%
	\title{Parallax Attention for Unsupervised Stereo Correspondence Learning}

	\author{Longguang Wang, Yulan Guo, Yingqian Wang, Zhengfa Liang, Zaiping Lin, Jungang Yang, Wei An
		\IEEEcompsocitemizethanks{\IEEEcompsocthanksitem Longguang Wang, Yulan Guo, Yingqian Wang, Zaiping Lin, Jungang Yang, and Wei An  are with the College of Electronic Science and Technology, Nation University of Defense Technology (NUDT), P. R. China. Yulan Guo is also with the School of Electronics and Communication Engineering, Sun Yat-sen University, Guangzhou, 510275, China.\protect\\
		E-mail:\{wanglongguang15, yulan.guo, wangyingqian16, linzaiping, yangjungang, anwei\}@nudt.edu.cn.\protect\\
		Zhengfa Liang is with the National Key Laboratory of Science and Technology on Blind Signal Processing, P. R. China. Email: liangzhengfa10@nudt.edu.cn.
		\IEEEcompsocthanksitem \textcolor{black}{L. Wang and Y. Guo have equal contribution to this work and are co-first authors.} Corresponding author: Yulan Guo.}
		}

	\markboth{IEEE Transactions on Pattern Analysis and Machine Intelligence,~Vol.~XX, No.~XX, XX~2020}%
	{Wang \MakeLowercase{\textit{et al.}}: Parallax Attention for Unsupervised Stereo Correspondence Learning}

	\IEEEtitleabstractindextext{%
		\begin{abstract}
			Stereo image pairs encode 3D scene cues into stereo correspondences between the left and right images. To exploit 3D cues within stereo images, recent CNN based methods commonly use cost volume techniques to capture stereo correspondence over large disparities. However, since disparities can vary significantly for stereo cameras with different baselines, focal lengths and resolutions, the fixed maximum disparity used in cost volume techniques hinders them to handle different stereo image pairs with large disparity variations. In this paper, we propose a generic parallax-attention mechanism (PAM) to capture stereo correspondence regardless of disparity variations. Our PAM integrates epipolar constraints with attention mechanism to calculate feature similarities along the epipolar line to capture stereo correspondence. Based on our PAM, we propose a parallax-attention stereo matching network (PASMnet) and a parallax-attention stereo image super-resolution network (PASSRnet) for stereo matching and stereo image super-resolution tasks. Moreover, we introduce a new and large-scale dataset named Flickr1024 for stereo image super-resolution. Experimental results show that our PAM is generic and can effectively learn stereo correspondence under large disparity variations in an unsupervised manner.
			Comparative results show that our PASMnet and PASSRnet achieve the state-of-the-art performance. 
		\end{abstract}
		
		}

	\maketitle

	\IEEEdisplaynontitleabstractindextext

	%
	\IEEEpeerreviewmaketitle

	\IEEEraisesectionheading{\section{Introduction}}
	
	\IEEEPARstart{W}{ith} the popularity of dual cameras in mobile phones, autonomous vehicles and robots, stereo vision has attracted increasingly attention in both academia and industry \cite{Guo20143D, Guo2020Deep}. Traditional studies \cite{Luo2016Efficient,Kendall2017End,Chang2018Pyramid} mainly focus on finding correspondences between stereo images to provide depth information, namely stereo matching. Recent works further use 3D cues within stereo images for various tasks including stereo image restoration \cite{Wang2019Learning,Zhou2019DAVANet}, stereo magnification \cite{Zhou2018Stereo}, stereo video retargeting \cite{Li2018Depth} and stereo neural style transfer \cite{Chen2018Stereoscopic}. In real-world applications such as mobile phones, the baselines and resolutions of stereo cameras in various devices are different. Moreover, 	
	the focal length of stereo cameras may change as the cameras adjust to a new scene. Since disparities between stereo images can vary significantly for stereo cameras with different baselines, focal lengths and resolutions, it is highly challenging to incorporate stereo correspondence for various real-world applications.
	
	Motivated by the powerful feature representation capability of convolutional neural network (CNN), deep feature representations are widely used to calculate similarities for stereo correspondence. However, the local receptive field hinders plain CNNs to capture correspondence over large disparities. To overcome this limitation, cost volumes are widely applied in CNN-based methods \cite{Mayer2016Large,Kendall2017End,Chang2018Pyramid,Jie2018Left}. Several methods \cite{Kendall2017End,Chang2018Pyramid,Guo2019Group} concatenate unary features from left and right images  to generate a 4D cost volume (\emph{i.e.}, height$\times$width$\times$disparity$\times$channel). Then, 3D CNNs are used to learn matching costs within the cost volume. However, learning matching costs from 4D cost volumes suffers from a high computational and memory burden. To achieve higher efficiency, 1D correlation is used to reduce feature dimension \cite{Mayer2016Large,Jie2018Left,Liang2018Learning,liang2019stereo}, resulting in 3D cost volumes (\emph{i.e.}, height$\times$width$\times$disparity). Although 4D/3D cost volumes enable CNNs to capture correspondence over large disparities, the fixed maximum disparity hinders them to handle different stereo image pairs with large disparity variations.
	
	In this paper, we propose an unsupervised parallax-attention mechanism (PAM) to learn  stereo correspondence under large disparity variations. Our PAM integrates epipolar constraints with attention mechanism to calculate feature similarities along the epipolar line. Specifically, for each pixel in the left image, its feature similarities with all possible disparities in the right image are computed to generate an attention map. By regularizing the attention map, our PAM can focus on the most similar feature to provide accurate stereo correspondence.
	
\begin{figure}[t]
		\centering
		\includegraphics[width=1\linewidth]{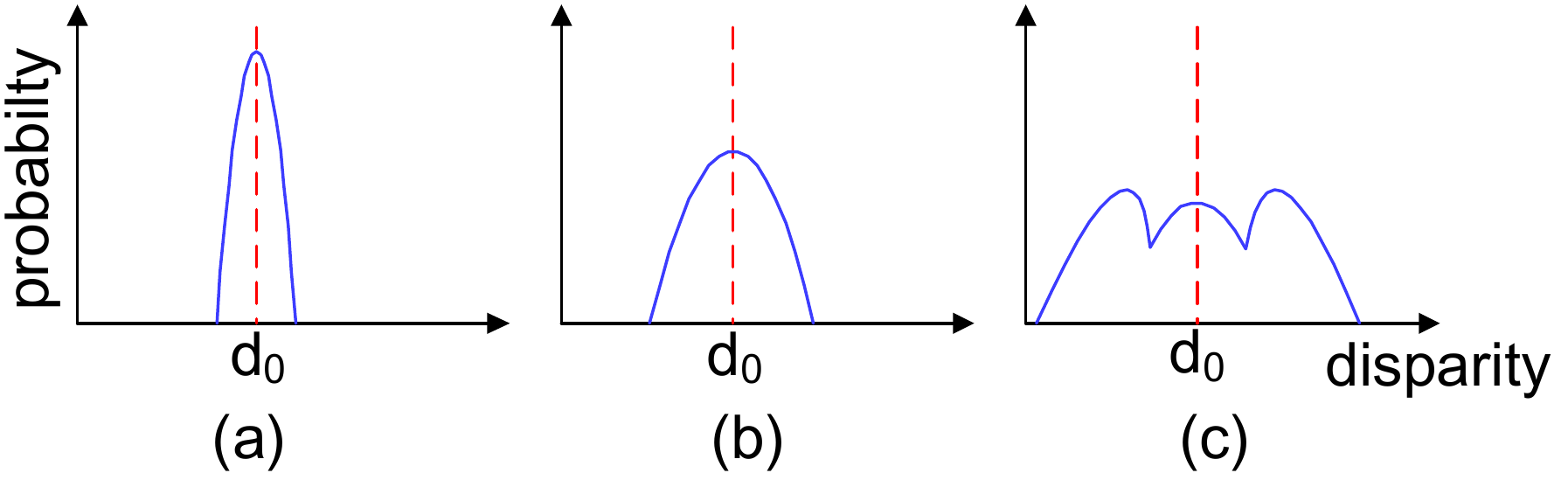}
		\caption{An illustration of the ambiguity for disparity regression in existing cost volume based methods. These three distributions have equal regression results $d_0$ but are quite different. A unimodal distribution peaked at the true disparity with high sharpness (\emph{i.e.}, low uncertainty) is more reasonable, as shown in (a).}
		\label{fig0}
	\end{figure}

	Compared to cost volume techniques, our PAM has three remarkable properties: \textbf{First}, the PAM gets access to different disparities through matrix multiplication instead of shift operation. Therefore, our PAM does not need to manually set a fixed maximum disparity and can handle large disparity variations. \textcolor{black}{\textbf{Second}, cost volume based methods commonly regress disparities upon matching costs and then calculate losses upon these disparities. However, this may lead to unreasonable cost distribution due to the ambiguity of disparity regression (as shown in Fig.~\ref{fig0}). In contrast, performing direct regularization on parallax-attention maps (\emph{e.g.}, through smoothness loss and cycle loss, as described in Sec.~\ref{Sec4.2}) enables our PAM to achieve improved performance.} \textbf{Third}, the PAM is compact and can be applied to various tasks such as stereo matching and stereo image super-resolution (SR). By using PAM, 3D cues from left and right images can be aggregated without explicit disparity calculation.

	This paper is an extension of our previous conference version \cite{Wang2019Learning}. The contributions of this work can be summarized as follows:
	\begin{itemize}
		\item A generic parallax-attention mechanism is proposed to learn stereo correspondence in image pairs with large disparity variations in an unsupervised manner.
		\item The PAM is successfully applied to two specific tasks: stereo matching and stereo image SR. Our PAM based networks achieve the state-of-the-art performance in both stereo matching and stereo image SR.
		\item A new dataset, namely Flickr1024, is proposed for the training of stereo image SR networks. This dataset consists of 1024 high-quality stereo image pairs and covers various scenes.
	\end{itemize}
	
The rest of this paper is organized as follows. In Section \ref{Sec2}, we briefly review the related work. In Section \ref{Sec3}, we present our PAM in details. In Sections \ref{Sec4} and \ref{Sec5}, we apply the PAM to stereo matching and stereo image SR tasks, respectively. Finally, we conclude this paper in Section \ref{Sec6}.
	
	\section{Related Work}
	\label{Sec2}
    Stereo vision has been investigated for various tasks in recent decades \cite{Luo2016Efficient,Chang2018Pyramid,Wang2019Learning,Zhou2018Stereo,Zhou2019DAVANet,Chen2018Stereoscopic}. Here, we focus on two specific tasks, including stereo matching and stereo image SR. Moreover, we also review attention mechanisms that are highly related to our work.  
	
	\subsection{Stereo Matching}
	\subsubsection{Supervised Stereo Matching}
	Traditional stereo matching methods commonly follow a popular four-step pipeline, including matching cost calculation, cost aggregation, disparity calculation and disparity refinement \cite{Liang2018Learning}. 
	
	Motivated by the success of CNN in many vision tasks such as object recognition \cite{He2015Spatial,He2016Deep} and detection \cite{Redmon2016You,He2017Mask}, some early learning-based methods \cite{Zbontar2015Computing,Zagoruyko2015Learning,Shaked2017Improved} use CNNs to replace one or multiple steps in the stereo matching pipeline. Zbontar and Lecun \cite{Zbontar2015Computing} used a CNN to compute matching cost between two image patches. The resulting matching costs are then processed by several traditional techniques (including cross-based cost aggregation and semi-global matching) to predict the disparity map. Luo \emph{et al.} \cite{Luo2016Efficient} proposed a dot product operation to compute correlation between unary feature representations of left and right images. Their network has lower computational complexity and achieves higher efficiency than \cite{Zbontar2015Computing}. To improve feature representation capability, pyramid pooling \cite{Park2017Look} and highway networks \cite{Shaked2017Improved} are used to enlarge the receptive field and deepen the network, respectively. These two methods produce more accurate disparities than \cite{Zbontar2015Computing}.
	
	To achieve better performance, several methods \cite{Mayer2016Large,Kendall2017End,Liang2018Learning} are proposed to integrate all steps into an end-to-end architecture for joint optimization. Mayer \cite{Mayer2016Large} proposed an end-to-end CNN called DispNet to regress disparities from stereo images. Specifically, 1D correlation operation is used to generate 3D cost volumes for matching cost learning. Following DispNet, several works \cite{Yang2018Segstereo,Liang2018Learning} have been proposed to employ correlation operation and 3D cost volumes for disparity estimation. Liang \emph{et al.} \cite{Liang2018Learning} used feature constancy to bridge the gap between disparity calculation and disparity refinement. Therefore, their network can seamlessly integrate the disparity calculation and disparity refinement steps into a compact network to improve efficiency.
	
	Recent methods \cite{Kendall2017End,Chang2018Pyramid,Guo2019Group} commonly use concatenation to generate 4D cost volumes and then use 3D CNNs for matching cost aggregation. GC-Net \cite{Kendall2017End} is the first work to use naive concatenation instead of correlation to build 4D cost volumes. Specifically, 3D convolutions are used to aggregate context information within the 4D cost volume for matching cost aggregation. Following GC-Net, Chang \emph{et al.} \cite{Chang2018Pyramid} proposed a pyramid stereo matching network (PSMNet) with a spatial pyramid pooling module to enlarge receptive fields for more representative features. In PSMNet, stacked 3D hourglass networks are used to aggregate matching costs at multiple scales. Inspired by the semi-global matching (SGM) method \cite{Hirschmuller2007Stereo}, Zhang \emph{et al.} \cite{Zhang2019GA} introduced semi-global and local guided aggregation layers to aggregate matching costs in different directions and local regions. 
	
	\subsubsection{Unsupervised Stereo Matching}
	Due to the difficulty of collecting a large dataset with densely labeled groundtruth depth, several methods \cite{Zhou2017Unsupervised,Yang2018Segstereo,Li2018Occlusion} have been developed to learn stereo matching in an unsupervised manner. 
	
	Zhou \emph{et al.} \cite{Zhou2017Unsupervised} adopted left-right consistency check to produce a confidence map to guide the training of a network. Yang \emph{et al.} \cite{Yang2018Segstereo} proposed a SegStereo network to use semantic cues for stereo matching. Specifically, they introduced a segmentation sub-network to provide semantic features and proposed a semantic loss regularization to improve the robustness of disparity estimation. Li \emph{et al.} \cite{Li2018Occlusion} proposed an occlusion aware stereo matching network (OASM) to exploit occlusion cues for stereo matching. Specifically, they introduced an occlusion inference module to provide occlusion cues and proposed a hybrid loss to use the interaction between disparity and occlusion as the supervision for network training. 
	
	Existing supervised and unsupervised stereo matching methods commonly used 3D/4D cost volumes to capture stereo correspondence over large disparities. However, the fixed maximum disparity hinders them to handle different stereo images with large disparity variations. Moreover, 4D cost volumes require high computational and memory costs.
	
	\subsection{Stereo Image Super-Resolution}
	Stereo image SR aims at recovering a high-resolution (HR) left image from a pair of low-resolution (LR) stereo images. Given additional information from a second viewpoint, stereo correspondence can be used to improve the SR performance. 
	\textcolor{black}{A recent work \cite{Song2020Stereoscopic} further extend the stereo image SR task to a stereoscopic image SR task that reconstructs a pair of HR left/right images while preserving stereo consistency. Note that, we only focus on the stereo image SR task in this paper.}
	Due to the unique characteristics of stereo images (such as epipolar constraints and non-local correspondence), video SR and light-field image SR methods are unsuitable for stereo image SR. Video SR methods \cite{Tao2017Detail,Wang2019EDVR,Wang2020Deep} commonly focus on the exploitation of local correspondence since motion between adjacent frames are mainly limited within a local region. Therefore, these video SR methods are unsuitable for stereo image SR due to the non-local correspondence over large disparities in stereo images. Light-field image SR methods \cite{Wang2018LFNet,Zhang2019Residuala,Wang2019Spatial} are specifically proposed for light-field images with short baselines. Therefore, they cannot be directly used for stereo image SR since stereo images usually have much longer baselines than light-field images.
	
	To enhance the resolution of stereo images, Bhavsar \emph{et al.} \cite{Bhavsar2010Resolution} argued that image SR and HR depth estimation are intertwined under stereo setting. They proposed a joint framework to iteratively estimate SR image and HR disparity. Recently, Jeon \emph{et al.} \cite{Jeon2018Enhancing} proposed to employ a parallax prior in CNN for stereo image SR. Given a stereo image pair, the right image is shifted with different intervals and concatenated with the left image to generate a stereo tensor. The stereo tensor is then fed to a CNN called StereoSR to generate SR results. However, StereoSR cannot handle different stereo images with large disparity variations since the number of shifted right images is fixed.
	
	
	
	\begin{figure}[t]
		\centering
		\includegraphics[width=1\linewidth]{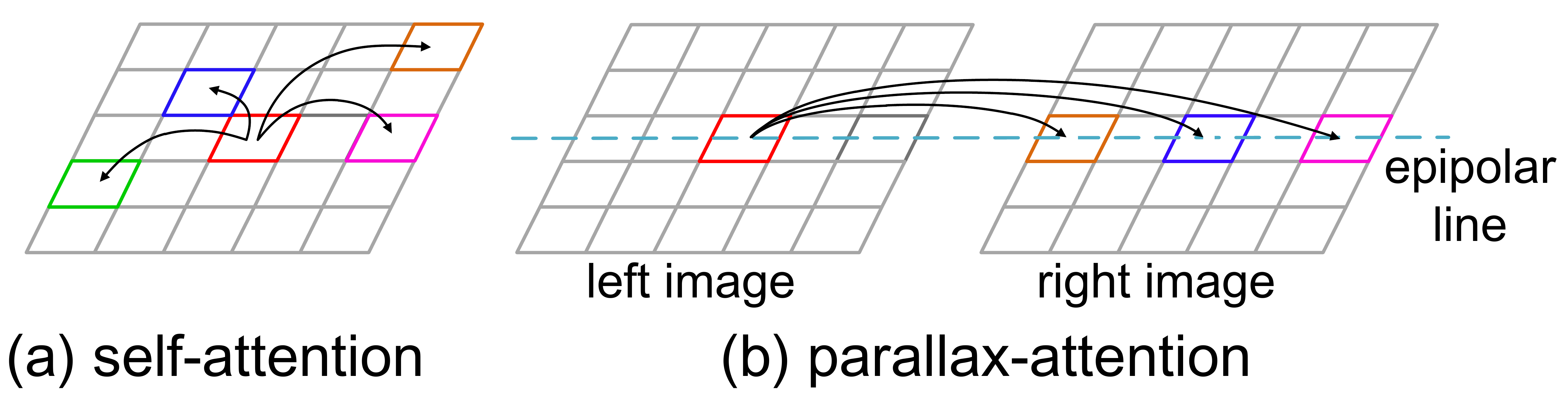}
		\caption{Comparison between self-attention mechanism and parallax-attention mechanism.}
		\label{fig1}
	\end{figure}
	\begin{figure}
		\centering
		\includegraphics[width=1\linewidth]{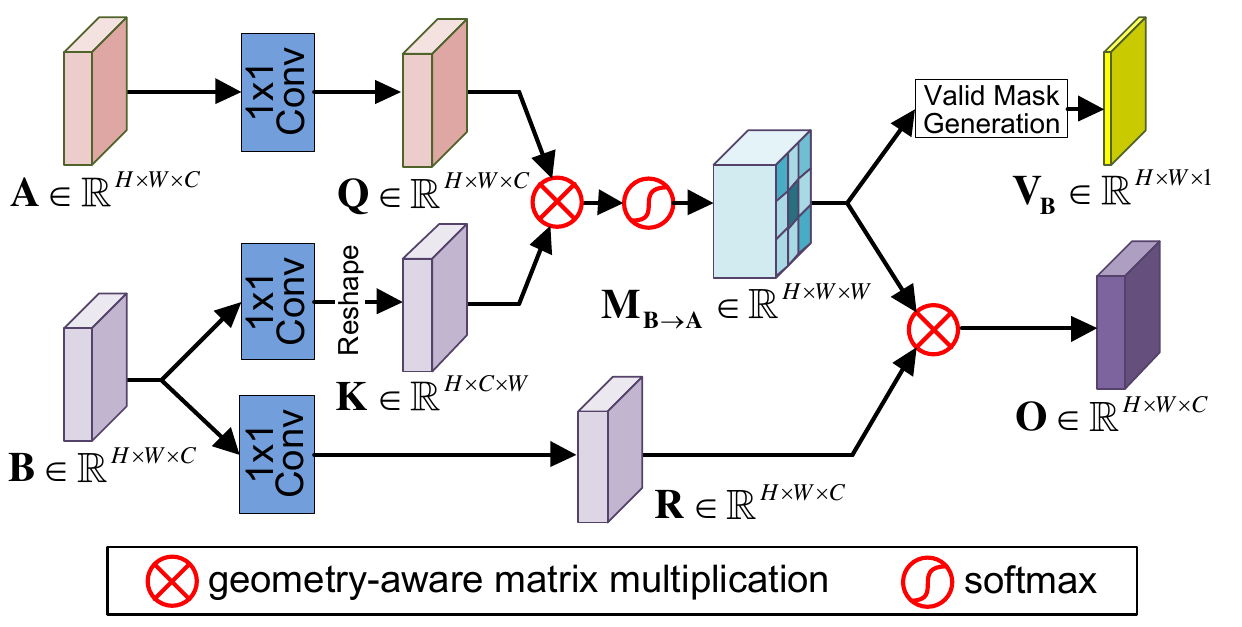}
		\caption{An illustration of our PAM.}
		\label{fig2}
	\end{figure}
	
	\subsection{Attention Mechanism}
	
	Attention mechanism was first introduced by Bahdanau \emph{et al.} \cite{Bahdanau2014Neural} and has been widely used in natural language processing tasks to capture long-range (long-term) dependencies \cite{Parikh2016Decomposable,Vaswani2017Attention}. Recently, attention mechanism has also been applied in many computer vision tasks including semantic segmentation \cite{Zhao2018PSANet,Fu2019Dual}, image captioning \cite{Xu2015Show,Chen2017SCA} and image generation \cite{Zhang2018Self}.
	Due to the limited receptive fields of CNNs, long-range dependencies cannot be captured. To overcome this limitation, self-attention mechanism is introduced to calculate the correlation of any two positions in an image.
	Wang \emph{et al.} \cite{Wang2018Non} proposed a non-local network to aggregate information from non-local regions, which can be viewed as a general format of the self-attention mechanism. 
	Zhang \emph{et al.} \cite{Zhang2018Self} introduced self-attention mechanism for image generation. The self-attention mechanism enables their network to use features in both local regions and distant regions to generate consistent scenarios. 
	Fu \emph{et al.} \cite{Fu2019Dual} proposed to use self-attention mechanism to aggregate long-range information for semantic segmentation. 
	
	
	Our work is motivated by the self-attention mechanism. We incorporate epipolar constraints with attention mechanism and develop a parallax-attention mechanism to capture correspondences in stereo images (as shown in Fig.~\ref{fig1}). For each pixel in the left image, our parallax-attention mechanism attends to all disparities along the epipolar line and learns to focus on the most similar one. 
	
	\section{Parallax-Attention Mechanism (PAM)}
	\label{Sec3}
	In this section, we present our PAM in details. We first introduce the formulation of our PAM, and then describe left-right consistency, cycle consistency and valid mask based on our PAM.
	
	\subsection{Formulation}
	
	\subsubsection{Overview}
	
	In self-attention mechanism, a feature map of size $\mathbb{R}^{{H}\times{W}\times{C}}$ is first reshaped to the size of $\mathbb{R}^{{HW}\times{C}}$. Then, matrix multiplication ($\mathbb{R}^{{HW}\times{C}}\times\mathbb{R}^{{C}\times{HW}}$) is used to calculate the correlation of any two positions in an image. For stereo images, the corresponding pixel for a pixel in the left image only lies along its epipolar line in the right image. Taking this epipolar constraint into consideration, our PAM uses specifically designed reshape operation and geometry-aware matrix multiplication to calculate the correlation between a pixel in the left image and all positions along the epipolar line in the right image. 
	
	As illustrated in Fig. \ref{fig2}, given two feature maps ${\rm\textbf{A}},{\rm\textbf{B}}\!\in\!\mathbb{R}^{{H}\times{W}\times{C}}$ from a stereo image pair, they are first fed to $1\times1$ convolutions for feature adaption. Specifically, ${\rm\textbf{A}}$ is fed to a $1\times1$ convolution to produce a query feature map ${\rm\textbf{Q}}\!\in\!\mathbb{R}^{{H}\times{W}\times{C}}$. 
	Meanwhile, ${\rm\textbf{B}}$ is fed to another $1\times1$ convolution to produce a key feature map ${\rm\textbf{K}}\!\in\!\mathbb{R}^{{H}\times{W}\times{C}}$, which is then reshaped to $\mathbb{R}^{{H}\times{C}\times{W}}$. Then, matrix multiplication is performed between $\rm\textbf{Q}$ and $\rm\textbf{K}$ and a softmax layer is applied, resulting in a parallax-attention map ${\rm\textbf{M}}_{{\rm\textbf{B}}\rightarrow{\rm\textbf{A}}}\!\in\!\mathbb{R}^{{H}\times{W}\times{W}}$. 
	Through matrix multiplication, our PAM can efficiently encode feature correlation between any two positions along the epipolar line into the parallax-attention map. The parallax-attention map has several remarkable properties, as described in Sec.~\ref{Sec3.2} and \ref{Sec3.3}.
	Next, $\rm\textbf{B}$ is fed to a $1\times1$ convolution to generate response feature map ${\rm\textbf{R}}\!\in\!\mathbb{R}^{{H}\times{W}\times{C}}$, which is further multiplied by ${\rm\textbf{M}}_{\rm\textbf{B}\rightarrow{\rm\textbf{A}}}$ to produce output features ${\rm\textbf{O}}\!\in\!\mathbb{R}^{{H}\times{W}\times{C}}$. Meanwhile, a valid mask ${\rm\textbf{V}}_{\rm\textbf{B}}$ is also generated (Section \ref{Sec3.3}).
	
	It should be noted that all disparities are considered in the PAM. That is, our PAM does not need to manually set a fixed maximum disparity and can 
	handle large disparity variations. Since PAM can learn to focus on the features at accurate disparities using feature similarities, correspondence can then be captured.
	
	\begin{figure}
		\centering
		\includegraphics[width=1\linewidth]{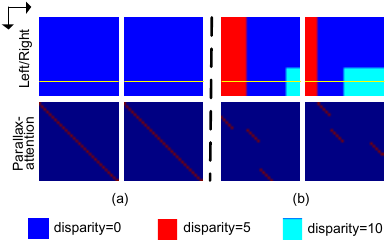}
		\caption{An illustration of our PAM using a toy example. The parallax-attention maps ($30\!\times\!30$) correspond to the regions ($1\!\times\!30$) marked by a yellow stroke. The first row represents left/right stereo images and the second row stands for parallax-attention maps $\rm\textbf{M}_{\emph{right}\rightarrow\emph{left}}$ and $\rm\textbf{M}_{\emph{left}\rightarrow\emph{right}}$.}
		\label{fig3}
	\end{figure}
	
	\subsubsection{Toy Example}
	\label{Sec3.1.2}
	We further illustrate our PAM using a toy example, as shown in Fig. \ref{fig3}. Given a stereo image pair ${\rm\textbf{I}}_\emph{left}$ and ${\rm\textbf{I}}_\emph{right}$ of size $\mathbb{R}^{{30}\times{30}}$, parallax-attention maps $\rm\textbf{M}_{\emph{left}\rightarrow\emph{right}}$ and $\rm\textbf{M}_{\emph{right}\rightarrow\emph{left}}$ of size $\mathbb{R}^{{30}\times{30}\times{30}}$ can be generated by our PAM. Note that, each slice of the parallax-attention maps (\emph{e.g.}, ${\rm\textbf{M}}_{\emph{right}\rightarrow\emph{left}}(i,:,:)$) delivers the dependency between corresponding rows (\emph{i.e.}, ${\rm\textbf{I}}_\emph{left}(i,:)$ and ${\rm\textbf{I}}_\emph{right}(i,:)$). 
	If the disparity between a pair of stereo images is zero, all parallax-attention maps are identity matrices, as shown in Fig. \ref{fig3} (a). That is because, the $j^{th}$ pixel in ${\rm\textbf{I}}_\emph{left}(i,:)$ corresponds to the $j^{th}$ pixel in ${\rm\textbf{I}}_\emph{right}(i,:)$. Therefore, position $(j,j)$ in ${\rm\textbf{M}}_{\emph{right}\rightarrow\emph{left}}$ is focused on. For regions with non-zero disparities, \emph{e.g.}, the red region in Fig. \ref{fig3} (b) with a disparity of 5, the $j^{th}$ pixel in ${\rm\textbf{I}}_\emph{left}(i,:)$ corresponds to the $(j\!-\!5)^{th}$ pixel in ${\rm\textbf{I}}_\emph{right}(i,:)$. Therefore, position $(j,j-5)$ in ${\rm\textbf{M}}_{\emph{right}\rightarrow\emph{left}}$ is focused on. In summary, stereo correspondence can be depicted by the positions of focused pixels in the parallax-attention maps. 
	
	Furthermore, occlusion can be encoded by the parallax-attention maps. Specifically, it can be observed from ${\rm\textbf{M}}_{\emph{right}\rightarrow\emph{left}}$ in Fig.~\ref{fig3}(b) that several vertical regions are ``discarded" without any position being focused on. That is because, these regions in ${\rm\textbf{I}}_\emph{right}(i,:)$ are occluded in ${\rm\textbf{I}}_\emph{left}(i,:)$, thus no correspondence should be focused on. Similar ``discarded" horizontal regions are also caused by occlusion.
	
	It should be noted that, only integer disparities are considered in our toy example, which is not the real case. In practice, our PAM can focus on adjacent pixels to handle sub-pixel disparities. Due to the softmax layer used in PAM, several pixels in ``discarded" horizontal regions may be incorrectly focused on. However, these occluded regions can be excluded using valid masks, as demonstrated in Sec.~\ref{Sec3.3}.
	
	\subsection{Left-Right Consistency and Cycle Consistency}
	\label{Sec3.2}
    To capture reliable and consistent correspondence, we introduce left-right consistency and cycle consistency to regularize our PAM. 
	
	Given feature representations extracted from a pair of stereo images ${\rm\textbf{I}}_\emph{left}$ and ${\rm\textbf{I}}_\emph{right}$, two parallax-attention maps (${\rm\textbf{M}}_{\emph{left}\rightarrow\emph{right}}$ and $\rm\textbf{M}_{\emph{right}\rightarrow\emph{left}}$) are generated by our PAM. Ideally, the following left-right consistency can be obtained if our PAM captures accurate correspondence:
	\begin{equation}
	\label{eq2}
	\left\{
	\begin{aligned}
	{\rm\textbf{I}}_\emph{left}&\!=\!{\rm\textbf{M}}_{\emph{right}\rightarrow\emph{left}}\otimes{\rm\textbf{I}}_\emph{right}\\
	{\rm\textbf{I}}_\emph{right}&\!=\!{\rm\textbf{M}}_{\emph{left}\rightarrow\emph{right}}\otimes{\rm\textbf{I}}_\emph{left}
	\end{aligned}
	\right.,
	\end{equation}
	where $\otimes$ denotes geometry-aware matrix multiplication. Based on Eq. (\ref{eq2}), we can further derive a cycle consistency:
	\begin{equation}
	\label{eq3}
	\left\{
	\begin{aligned}
	{\rm\textbf{I}}_\emph{left}&\!=\!{\rm\textbf{M}}_{\emph{left}\rightarrow\emph{right}\rightarrow\emph{left}}\otimes{\rm\textbf{I}}_\emph{left}\\
	{\rm\textbf{I}}_\emph{right}&\!=\!{\rm\textbf{M}}_{\emph{right}\rightarrow\emph{left}\rightarrow\emph{right}}\otimes{\rm\textbf{I}}_\emph{right}
	\end{aligned}
	\right.,
	\end{equation}
	where the cycle-attention maps $\rm\textbf{M}_{\emph{left}\rightarrow\emph{right}\rightarrow\emph{left}}$ and $\rm\textbf{M}_{\emph{right}\rightarrow\emph{left}\rightarrow\emph{right}}$ can be calculated as:
	\begin{equation}
	\left\{
	\begin{aligned}
	\rm\textbf{M}_{\emph{left}\rightarrow\emph{right}\rightarrow\emph{left}}&\!=\!\rm\textbf{M}_{\emph{right}\rightarrow\emph{left}}\otimes\rm\textbf{M}_{\emph{left}\rightarrow\emph{right}}\\
	\rm\textbf{M}_{\emph{right}\rightarrow\emph{left}\rightarrow\emph{right}}&\!=
	\!\rm\textbf{M}_{\emph{left}\rightarrow\emph{right}}\otimes\rm\textbf{M}_{\emph{right}\rightarrow\emph{left}}
	\end{aligned}
	\right..
	\end{equation}
	
	\begin{figure}[t]
		\centering
		\includegraphics[width=1\linewidth]{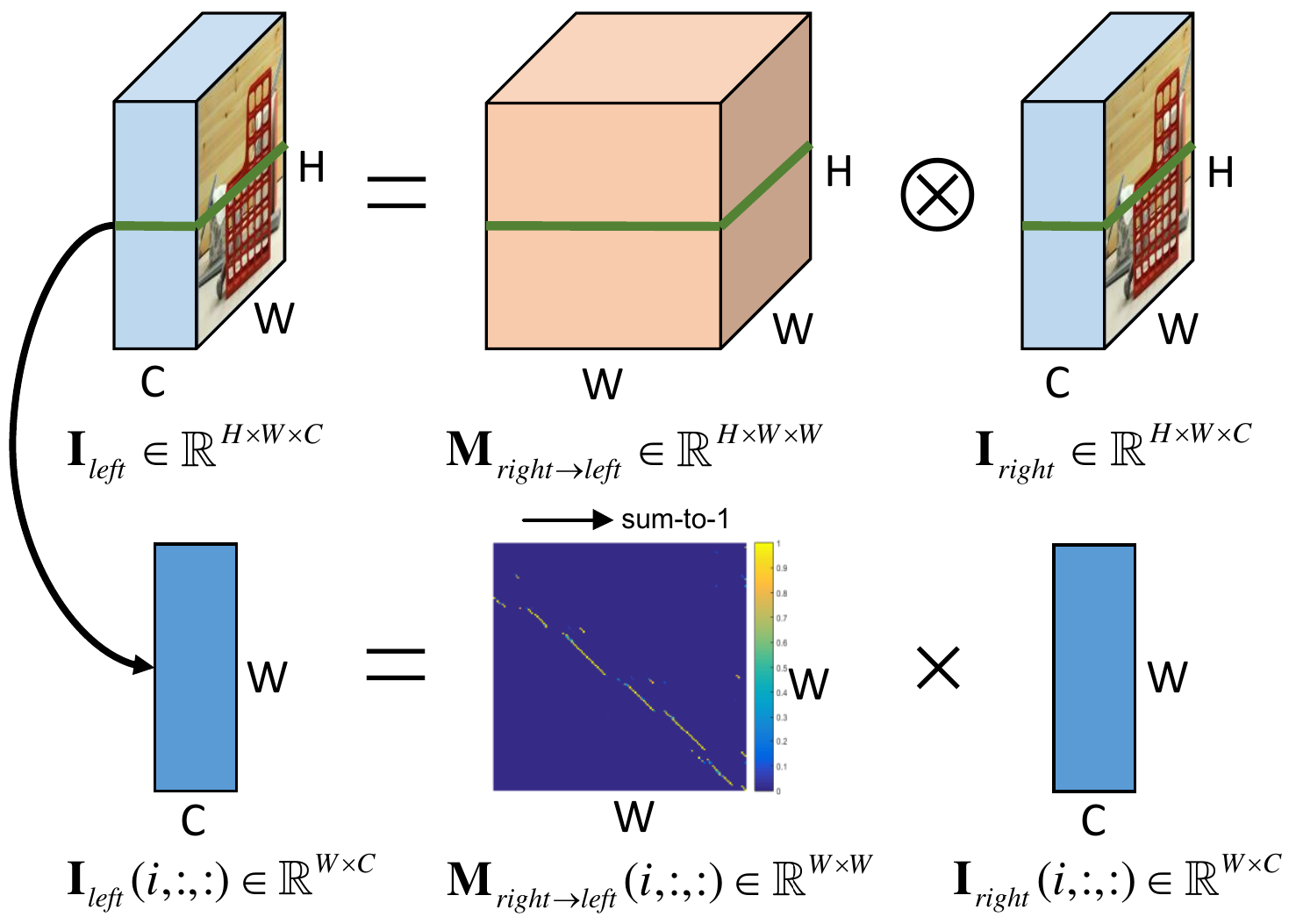}
		\caption{An illustration of geometry-aware matrix multiplication $\otimes$.}
		\label{fig4}
	\end{figure}
	
	We further illustrate geometry-aware matrix multiplication $\otimes$~\footnote{$\otimes$ can be implemented using tf.matmul() or torch.matmul().} in Fig. \ref{fig4}. Take Eq.~(1) for an example, the product of corresponding slices in $\rm\textbf{M}_{\emph{right}\rightarrow\emph{left}}$ and ${\rm\textbf{I}}_\emph{right}$ (\emph{e.g.}, $\rm\textbf{M}_{\emph{right}\rightarrow\emph{left}}$($i,:,:)\!\in\!\mathbb{R}^{\emph{W}\times\emph{W}}$ and ${\rm\textbf{I}}_\emph{right}$($i,:,:)\!\in\!\mathbb{R}^{\emph{W}\times\emph{C}}$) determines the $i^\textit{th}$ slice of ${\rm\textbf{I}}_\emph{left}$, \textit{i.e.}, ${\rm\textbf{I}}_\emph{left}$($i,:,:)\!\in\!\mathbb{R}^{\emph{W}\times\emph{C}}$. \textcolor{black}{All these slices are concatenated to obtain ${\rm\textbf{I}}_\emph{left}\!\in\!\mathbb{R}^{\emph{H}\times\emph{W}\times\emph{C}}$}.

	\subsection{Valid Mask}
	\label{Sec3.3}
	Since left-right consistency and cycle consistency do not hold for occluded regions, we perform occlusion detection based on the parallax-attention map to generate valid masks and only enforce consistency regularization on those valid regions.
	
\begin{figure}[t]
		\centering
		\includegraphics[width=0.9\linewidth]{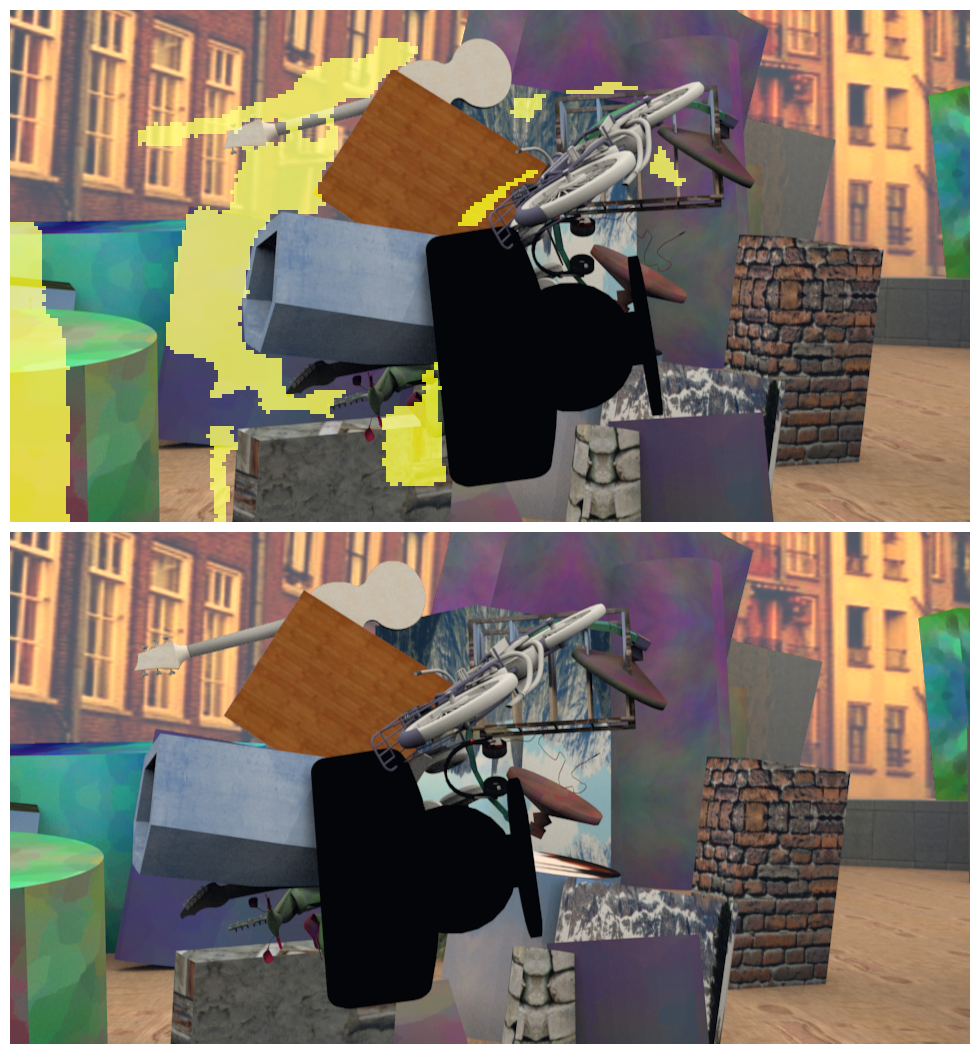}
		\caption{Visualization of valid masks. A stereo image pair and the occluded regions (\emph{i.e.}, yellow regions) are illustrated.}
		\label{fig5}
	\end{figure}

	As illustrated in the Sec.~\ref{Sec3.1.2}, occlusion is encoded as the ``discarded" regions in the parallax-attention maps. In practice, vertical ``discarded" regions that correspond to occluded regions are usually assigned with small weights in the parallax-attention maps (\emph{e.g.}, $\rm\textbf{M}_{\emph{left}\rightarrow\emph{right}}$). That is because, occluded pixels in the left image cannot find their correspondences in the right image, their feature similarities with all disparities are low. Therefore, a valid mask ${\rm\textbf{V}}_{\emph{left}}\!\in\!\mathbb{R}^{{H}\times{W}}$ can be obtained by:
	
	\begin{equation}
	{\rm\textbf{V}}_{\emph{left}}(i,k)\!=\!
	\left\{
	\begin{aligned}
	1,~~~&{\rm{if}}  \sum_{{j}\in[1,\,{W}]}{{\rm\textbf{M}}_{\emph{left}\rightarrow\emph{right}}(i,j,k)}>\textcolor{black}{\tau} \\
	0,~~~&\rm{otherwise}&
	\end{aligned},
	\right.
	\end{equation}
	where $\textcolor{black}{\tau}$ is a threshold (empirically set to 0.1 in this paper). An examples of valid mask is shown in Fig. \ref{fig5}.  In practice, we use several morphological operations to handle isolated pixels and holes in valid masks.
	
	In summary, our PAM provides a flexible and effective approach for unsupervised stereo correspondence learning and can be applied to various stereo vision tasks. In this paper, we demonstrate the effectiveness of PAM on two typical stereo tasks: stereo matching and stereo image SR.
	
	\begin{figure*}[t]
		\centering
		\includegraphics[width=1\linewidth]{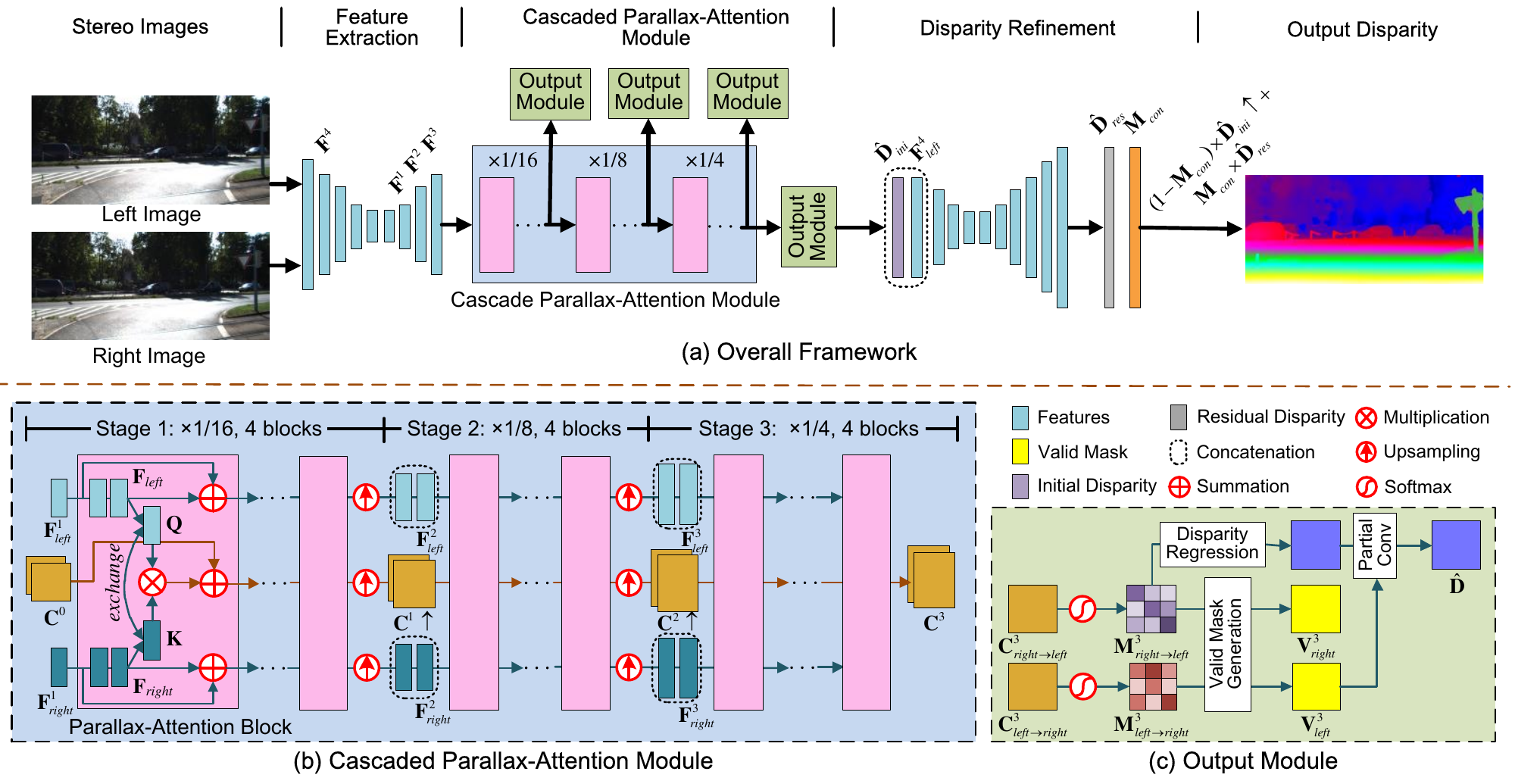}
		\caption{An overview of our PASMnet.}
		\label{fig11}
	\end{figure*}
	
	\begin{figure}[t]
		\centering
		\includegraphics[width=1\linewidth]{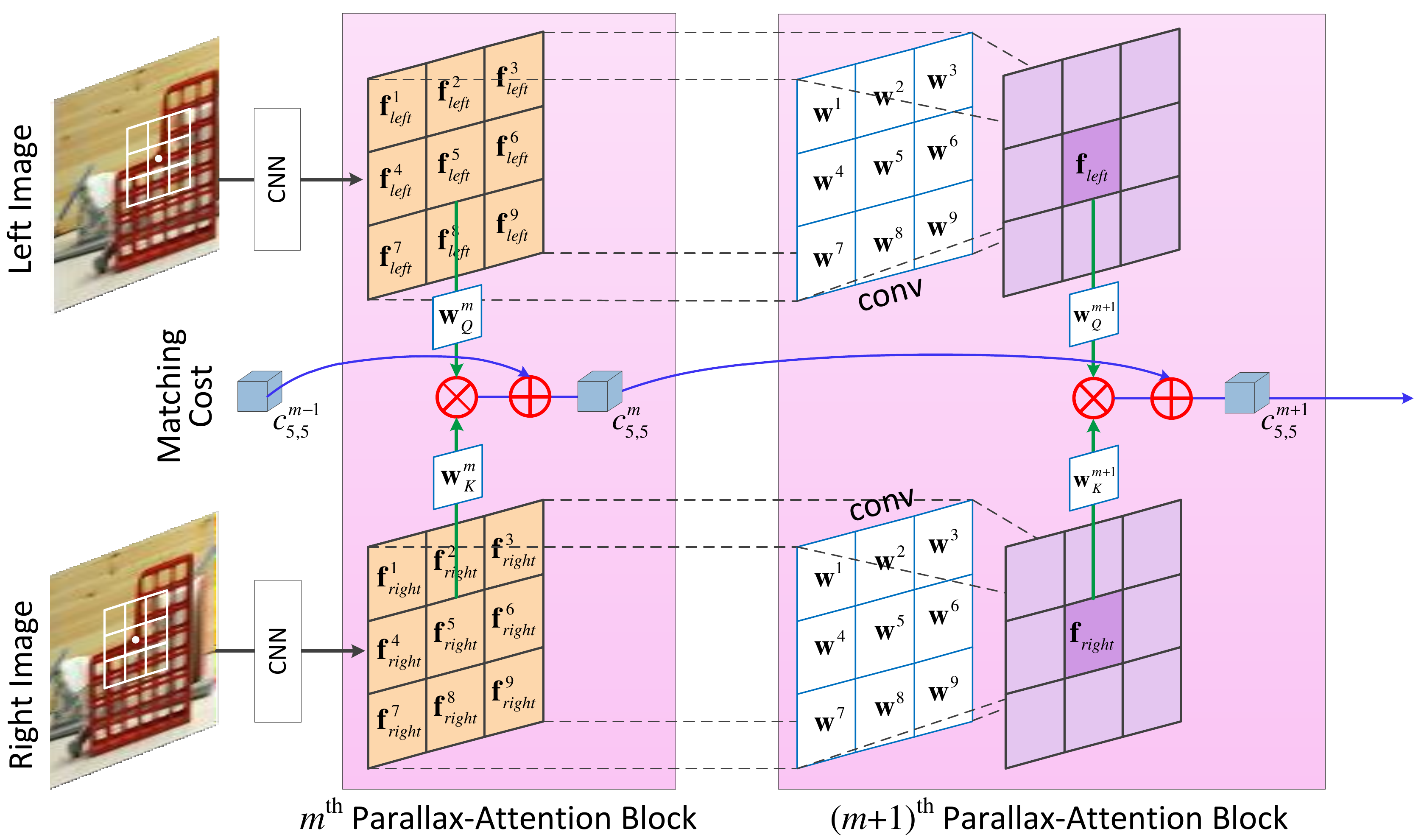}
		\caption{\textcolor{black}{An illustration of matching cost aggregation in our cascaded parallax-attention module.}}
		\label{fig14}
	\end{figure}
	
	\section{PAM for Unsupervised Stereo Matching}
	\label{Sec4}
    Stereo matching aims at finding correspondence pixels between a stereo image pair. Since disparities can vary significantly in real-world applications, the fixed maximum disparity in cost volume techniques hinders them to handle large disparity variations. In contrast, our PAM can get rid of setting a fixed maximum disparity by efficiently calculating feature correlation between any two positions along the epipolar line. Therefore, the PAM can be used for unsupervised stereo matching to handle large disparity variations. 
	
	\subsection{Network Architecture}
	\subsubsection{Overview}
	Based on our PAM, we proposed a parallax-attention stereo matching network (PASMnet). The architecture of our PASMnet is shown in Fig.~\ref{fig11}. Given a stereo image pair of size ${H}\times{W}$, they are first fed to an hourglass network for feature extraction. Then, the features extracted from left and right images are fed to a cascaded parallax-attention module to regress matching costs in a coarse-to-fine manner. Next, an initial disparity can be obtained from the matching cost using an output module. Finally, the initial disparity is further refined using an hourglass network to produce the output disparity.
	
	\subsubsection{Cascaded Parallax-Attention Module}
	After feature extraction, features from left and right images are fed to the cascaded parallax-attention module to regress matching costs in a coarse-to-fine manner. Specifically, our cascaded parallax-attention module consists of 3 stages (with 4 parallax-attention blocks in each stage), as shown in Fig.~\ref{fig11}(b). First, features from $\rm\textbf{F}^{1}$ (\emph{i.e.}, ${\rm\textbf{F}}^{1}_\emph{left},{\rm\textbf{F}}^\emph{1}_\emph{right}\in\mathbb{R}^{{\frac{H}{16}}\times{\frac{W}{16}}\times{C}}$) and  \textcolor{black}{initial matching costs ${\rm\textbf{C}}^{0}_{\emph{right}\rightarrow\emph{left}},{\rm\textbf{C}}^{0}_{\emph{left}\rightarrow\emph{right}}\in\mathbb{R}^{{\frac{H}{16}}\times{\frac{W}{16}}\times{\frac{W}{16}}}$} with all elements being set to 0 are passed to the first parallax-attention block. ${\rm\textbf{F}}^{1}_\emph{left}$ and ${\rm\textbf{F}}^{1}_\emph{right}$ are fed to two $3\times3$ convolutions to obtain ${\rm\textbf{F}}_\emph{left}$ and ${\rm\textbf{F}}_\emph{right}$, respectively. \textcolor{black}{Note that, the convolutions for left and right images share the same parameters.} Then, query features $\rm\textbf{Q}$ and key features $\rm\textbf{K}$ are obtained from ${\rm\textbf{F}}_\emph{left}$ and ${\rm\textbf{F}}_\emph{right}$ through $1\times1$ convolutions. Next, $\rm\textbf{K}$ is reshaped and multiplied with $\rm\textbf{Q}$, resulting in a matching cost ${\rm\textbf{C}}^1_{\emph{right}\rightarrow\emph{left}}\!\in\!\mathbb{R}^{{\frac{H}{16}}\times{\frac{W}{16}}\times{\frac{W}{16}}}$. Once ${\rm\textbf{C}}^1_{\emph{right}\rightarrow\emph{left}}$ is ready, ${\rm\textbf{F}}_\emph{left}$ and ${\rm\textbf{F}}_\emph{right}$ are exchanged to obtain ${\rm\textbf{C}}^1_{\emph{left}\rightarrow\emph{right}}$.
	\textcolor{black}{After that, ${\rm\textbf{C}}^0_{\emph{right}\rightarrow\emph{left}}$ and ${\rm\textbf{C}}^0_{\emph{left}\rightarrow\emph{right}}$ are added to ${\rm\textbf{C}}^1_{\emph{right}\rightarrow\emph{left}}$ and ${\rm\textbf{C}}^1_{\emph{left}\rightarrow\emph{right}}$, respectively.}
	Meanwhile, ${\rm\textbf{F}}^{1}_\emph{left}$ and ${\rm\textbf{F}}^{1}_\emph{right}$ are added to ${\rm\textbf{F}}_\emph{left}$ and ${\rm\textbf{F}}_\emph{right}$, respectively.

	As shown in Fig.~\ref{fig11}(b), the features and matching cost produced by the preceding parallax-attention block are fed to the succeeding block. The features produced by stage 1 are bilinearly upsampled and concatenated with $\rm\textbf{F}$$^{2}_\emph{left}$ and $\rm\textbf{F}$$^{2}_\emph{right}$ (from $\rm\textbf{F}$$^{2}$ in Fig.~\ref{fig11}(a)). Then, the concatenated features and the upsampled matching cost \textcolor{black}{($\rm\textbf{C}^1_{\emph{right}\rightarrow\emph{left}}\!\uparrow$,$\rm\textbf{C}_{\emph{left}\rightarrow\emph{right}}^1\!\uparrow$)} are passed to \textcolor{black}{stages} 2 and 3 for further refinement, resulting in the final matching cost \textcolor{black}{$\rm\textbf{C}^3_{\emph{right}\rightarrow\emph{left}},\rm\textbf{C}^3_{\emph{left}\rightarrow\emph{right}}\in\mathbb{R}^{{\frac{H}{4}}\times{\frac{W}{4}}\times{\frac{W}{4}}}$}.
	
	\textcolor{black}{Different from the cost volume based methods \cite{Kendall2017End,Chang2018Pyramid} that explicitly aggregate matching costs using 3D convolutions, our network performs implicit matching cost aggregation by cascading several parallax-attention blocks, as shown in Fig.~\ref{fig14}. Here, only one $3\times3$ convolutional layer (without batch normalization and ReLU layers) in the parallax-attention blocks is considered for simplicity. 
	In the $m^{\rm th}$ parallax-attention block, the matching cost $c^m_{5,5}$ \textcolor{black}{between the locations of ${\textbf{f}}^5_\emph{left}$ and ${\textbf{f}}^5_\emph{right}$} is computed as:
	\begin{equation}
	\begin{aligned}
		c_{5,5}^{m}
		&=c^{m-1}_{5,5}+({\rm \textbf{w}}_Q^m{\rm \textbf{f}}_\emph{left}^5)^{\rm T}{\rm \textbf{w}}_K^m{\rm \textbf{f}}_\emph{right}^5\\
		&=c^{m-1}_{5,5}+({\rm \textbf{f}}_\emph{left}^5)^{\rm T}({\rm \textbf{w}}_Q^m)^{\rm T}{\rm \textbf{w}}_K^m{\rm \textbf{f}}_\emph{right}^5\\
		&=c^{m-1}_{5,5}+\Psi^m\big({\rm \textbf{f}}_\emph{left}^5,{\rm \textbf{f}}_\emph{right}^5\big)
	\end{aligned}~, 
	\end{equation}
	where ${\rm \textbf{w}}_Q^m,{\rm \textbf{w}}_K^m\!\in\!\mathbb{R}^{\emph{\rm C}\times\emph{\rm C}}$ \textcolor{black}{are the kernels of two $1\times1$ convolutional layers}, ${\rm \textbf{f}}^5_\emph{left},{\rm \textbf{f}}^5_\emph{right}\!\in\!\mathbb{R}^{\emph{\rm C}\times\emph{\rm 1}}$ \textcolor{black}{are the corresponding features}, and $\Psi^m(\cdot)$ represents the matching cost calculation between the pair of input features.
	In the $(m+1)^{\rm th}$ block, features from the $m^{\rm th}$ block within a local neighborhood are aggregated to produce ${\rm \textbf{f}}_\emph{left}$ and ${\rm \textbf{f}}_\emph{right}$:
	\begin{equation}
	\left\{
	\begin{aligned}
		{\rm \textbf{f}}_\emph{left}&=\sum\nolimits_{i=1}^{9}{{\rm \textbf{w}}^i{\rm \textbf{f}}_\emph{left}^i}  \\
		{\rm \textbf{f}}_\emph{right}&=\sum\nolimits_{j=1}^{9}{{\rm \textbf{w}}^j{\rm \textbf{f}}_\emph{right}^j} \\
	\end{aligned}
	\right.,
	\end{equation}
	where ${\rm \textbf{w}}^i,{\rm \textbf{w}}^j\!\in\!\mathbb{R}^{\emph{\rm C}\times\emph{\rm C}}$ \textcolor{black}{are kernels at different sub-locations for two $3\times3$ convolutional layers with shared parameters}, and ${\rm \textbf{f}}_\emph{left},{\rm \textbf{f}}_\emph{right}\!\in\!\mathbb{R}^{\emph{\rm C}\times\emph{\rm 1}}$ \textcolor{black}{represent the output features}.
	Consequently, the matching cost $c^{m+1}_{5,5}$ is computed as:
	\begin{equation}
	\begin{aligned}
	c_{5,5}^{m+1}
	&=c^m_{5,5}\!+\!({\rm \textbf{w}}_Q^{m+1}{\rm \textbf{f}}_\emph{left})^{\rm T}{\rm \textbf{w}}_K^{m+1}{\rm \textbf{f}}_\emph{right}\\
	&=c^m_{5,5}\!+\!\sum\nolimits_i\sum\nolimits_j({\rm \textbf{w}}_Q^{m+1}{\rm \textbf{w}}^i{\rm \textbf{f}}_\emph{left}^i)^{\rm T}{\rm \textbf{w}}_K^{m+1}{\rm \textbf{w}}^j{\rm \textbf{f}}_\emph{right}^j\\
	&=c^m_{5,5}\!+\!\sum\nolimits_i\sum\nolimits_j({\rm \textbf{f}}_\emph{left}^i)^{\rm T}({\rm \textbf{w}}^i)^{\rm T}({\rm \textbf{w}}_Q^{m+1})^{\rm T}{\rm \textbf{w}}_K^{m+1}{\rm \textbf{w}}^j{\rm \textbf{f}}_\emph{right}^j\\
	&=c^m_{5,5}\!+\!\sum\nolimits_i\sum\nolimits_j\Psi^{m+1}_{i,j}({\rm \textbf{f}}_\emph{left}^i,{\rm \textbf{f}}_\emph{right}^j)
	\end{aligned},
	\label{eq7}
	\end{equation}
	where ${\rm \textbf{w}}_Q^{m+1},{\rm \textbf{w}}_K^{m+1}\!\in\!\mathbb{R}^{\emph{\rm C}\times\emph{\rm C}}$ \textcolor{black}{are the kernels of two $1\times1$ convolutional layers}, and $\Psi^{m+1}_{i,j}(\cdot)$ represents the matching cost calculation between the pair of input features. \textcolor{black}{Note that, ${\rm \textbf{f}}_\emph{left}^i$ and ${\rm \textbf{f}}_\emph{right}^j$ are from the $m^{\rm th}$ parallax-attention block. That is,} matching costs in a local neighborhood are implicitly aggregated as parallax-attention blocks are cascaded. Moreover, residual connection further enables the aggregation of matching costs from different network depths.   
	}
	
	\subsubsection{Output Module}
	As shown in Fig.~\ref{fig11}(a), the matching cost of the last block at each stage in the cascaded parallax-attention module is fed to an output module. \textcolor{black}{Within the output module of stage 3, $\rm{\textbf{C}}$$^3_{\emph{right}\rightarrow\emph{left}}$ and $\rm{\textbf{C}}$$^3_{\emph{left}\rightarrow\emph{right}}$ are first fed to a softmax layer to produce parallax-attention maps ${\rm\textbf{M}}^3_{\emph{right}\rightarrow\emph{left}}$ and $ {\rm\textbf{M}}^3_{\emph{left}\rightarrow\emph{right}}$$\in\mathbb{R}^{{\frac{H}{4}}\times{\frac{W}{4}}\times{\frac{W}{4}}}$, as shown in Fig.~\ref{fig11}(c).} Next, ${\rm\textbf{M}}^3_{\emph{right}\rightarrow\emph{left}}$ and ${\rm\textbf{M}}^3_{\emph{left}\rightarrow\emph{right}}$ are used to generate valid masks ${\rm\textbf{V}}^3_\emph{right}$ and ${\rm\textbf{V}}^3_\emph{left}$. Once ${\rm\textbf{M}}^3_{\emph{right}\rightarrow\emph{left}}$ is obtained, the disparity can be regressed as:
	
	\begin{equation}
	{\rm{\hat{\textbf{D}}}}(i,j)=\sum_{k=0}^{W/4-1}{(j-k)\times{\rm\textbf{M}}^3_{\emph{right}\rightarrow\emph{left}}(:,:,k)}.
	\end{equation}
	
	Note that, the estimated disparity is a sum of all disparity candidates weighted by the parallax-attention map. Consequently, our PASMnet does not need to manually set a fixed maximum disparity and can handle large disparity variations. Since pixels in occluded regions cannot find their correspondences, disparity in occluded regions cannot be well handled. Therefore, we exclude these invalid disparities and fill occluded regions using partial convolution \cite{Liu2018Image}.
	
	\subsubsection{Disparity Refinement}
	After the cascaded parallax-attention module, a disparity refinement module is introduced using features from the left image as guidance to provide structural information like edges. As shown in Fig.~\ref{fig11}(a), the initial disparity ${\rm\hat{\textbf{D}}}_\emph{ini}$ is concatenated with ${\rm\textbf{F}}^4_\emph{left}$ and fed to an hourglass network to produce a residual disparity map ${\rm\hat{\textbf{D}}}_\emph{res}$ and a confidence map ${\rm{\textbf{M}}}_\emph{con}$. Finally, the refined disparity is calculated as:
	\begin{equation}
	{\rm{\hat{\textbf{D}}}}_\emph{refined}=(1-{\rm{\textbf{M}}}_\emph{con})\times{\rm\hat{\textbf{D}}}_\emph{ini}\uparrow+{\rm{\textbf{M}}}_\emph{con}\times{\rm\hat{\textbf{D}}}_\emph{res},
	\end{equation}
	where $\uparrow$ is a bilinear upsampling operator.

	\subsection{Losses}
	\label{Sec4.2}
	\subsubsection{Photometric Loss}
	Following \cite{Yin2018GeoNet,Godard2017Unsupervised,Li2018Occlusion}, we introduce a photometric loss consisting of a mean absolute error (MAE) loss term and a structural similarity index (SSIM) loss term. Note that, since photometric consistency only holds in non-occluded regions, the photometric loss is formulated as:
	\begin{equation}
	\mathcal{L}_{\rm{p}}\!=\!\frac{1}{N}\sum_{p\in{{\rm\textbf{V}}_\emph{left}}}\!
	\alpha\frac{1-\mathcal{S}({\rm\textbf{I}}_\emph{left}(p),{\rm\hat{\textbf{I}}}(p))}{2}+
	(1-\alpha)\left\Vert{\rm\textbf{I}}_\emph{left}(p)\!-{\rm\hat{\textbf{I}}}(p)\right\Vert_{1},
	\end{equation}
	where ${\rm\hat{\textbf{I}}}=\mathcal{W}({\rm\textbf{I}}_\emph{right},\hat{\rm\textbf{D}}_\emph{refined})$ and $\mathcal{W}$ is a warping operator using the refined disparity. $\mathcal{S}$ is an SSIM function, $p$ represents a valid pixel covered by the valid mask, $N$ is the number of valid pixels, and $\alpha$ is empirically set to 0.85 in this paper.
	
	\subsubsection{Smoothness Loss}
	Following \cite{Yin2018GeoNet,Li2018Occlusion}, we use an edge-aware smoothness loss to encourage local smoothness of the disparity, which is defined as:
	\begin{equation}
	\begin{aligned}
	\mathcal{L}_{\mathrm{s}}\!=\!\frac{1}{N}\!\sum_{p}\!
	(&\left\Vert{\nabla_x{{\rm{\hat{\textbf{D}}}}_\emph{refined}}(p)}\right\Vert_{1}\!e^{\!-\!\left\Vert{\nabla_x{\rm\textbf{I}}_\emph{left}(p)}\right\Vert_{1}}+\\
	&\left\Vert{\nabla_y{{\rm{\hat{\textbf{D}}}}_\emph{refined}}(p)}\right\Vert_{1}\!e^{\!-\!\left\Vert{\nabla_y{\rm\textbf{I}}_\emph{left}(p)}\right\Vert_{1}}),
	\end{aligned}
	\end{equation}
	where $\nabla_x$ and $\nabla_y$ are gradients in the $x$ and $y$ axes, respectively. 
	
	\subsubsection{PAM Loss}
	\label{Sec4.2.3}
	We introduce three additional losses to regularize our PAM at multiple scales to capture accurate and consistent stereo correspondence. The PAM loss term for scale $s (s=1,2,3)$ is defined as:
	\begin{equation}
	\mathcal{L}^{s}_{\rm{PAM}}=\mathcal{L}^{s}_{\rm{PAM\emph{-}p}}+\lambda_{\rm{PAM\emph{-}s}}\mathcal{L}^{s}_{\rm{PAM\emph{-}s}}+\lambda_{\rm{PAM\emph{-}c}}\mathcal{L}^{s}_{\rm{PAM\emph{-}c}}.
	\end{equation}
	
	\noindent$\bullet$ {$\mathcal{L}^{s}_{\rm{PAM\emph{-}p}}$}
	\vspace{1.5mm}
	
	Different from the aforementioned photometric loss, here, we introduce a photometric loss based on the parallax-attention maps as:
	\begin{equation}
	\begin{aligned}
	\mathcal{L}^{s}_{\mathrm{PAM\emph{-}p}}=&\frac{1}{N^{s}_\emph{left}}\sum_{p\in{\rm\textbf{V}}^{s}_\emph{left}}
	\left\Vert{\rm\textbf{I}}^{s}_\emph{left}(p)-({\rm\textbf{M}}^{s}_{\emph{right}\rightarrow\emph{left}}\otimes{\rm\textbf{I}}^{s}_\emph{right})(p)\right\Vert_{1}+\\
	&\frac{1}{N^{s}_\emph{right}}\sum_{p\in{\rm\textbf{V}}^{s}_\emph{right}}
	\left\Vert {\rm\textbf{I}}^{s}_\emph{right}(p)\!-\!({\rm\textbf{M}}^{s}_{\emph{left}\rightarrow\emph{right}}\otimes{\rm\textbf{I}}^{s}_\emph{left})(p)\right\Vert_{1}.
	\end{aligned}
	\end{equation}
	where $N^{s}_\emph{left}$ and $N^{s}_\emph{right}$ are the numbers of valid pixels in ${\rm\textbf{V}}^{s}_\emph{left}$ and ${\rm\textbf{V}}^{s}_\emph{right}$, respectively. ${\rm\textbf{I}}^{s}_\emph{left}$ and ${\rm\textbf{I}}^{s}_\emph{right}$ are bilinearly downsampled images at corresponding scale level.
	
	\noindent$\bullet$ {$\mathcal{L}^{s}_{\rm{PAM\emph{-}s}}$}
	\vspace{1.5mm}

	Different from the aforementioned smoothness loss, here, another smoothness loss is introduced to directly regularize parallax-attention maps:
	
	\begin{equation}
	\begin{aligned}
	\mathcal{L}^{s}_{\rm{PAM\emph{-}s}}\!=\!\frac{1}{N^s}\sum_{\rm\textbf{M}}\sum_{i,j,k}(&\left\Vert {\rm\textbf{M}}^{s}(i,j,k)\!-\!{\rm\textbf{M}}^{s}(i\!+\!1,j,k)\right\Vert_{1}\!+\!\\
	&\left\Vert {\rm\textbf{M}}^{s}(i,j,k)-\!{\rm\textbf{M}}^{s}(i,j\!+\!1,k\!+\!1)\right\Vert_{1}),
	\end{aligned}
	\label{eq8}
	\end{equation}
	where ${\rm\textbf{M}}^{s}\!\in\!\left\{{\rm\textbf{M}}^{s}_{\emph{left}\rightarrow\emph{right}},{\rm\textbf{M}}^{s}_{\emph{right}\rightarrow\emph{left}}\right\}$ and $N^s$ is the number of pixels in ${\rm\textbf{M}}^{s}_{\emph{left}\rightarrow\emph{right}}$. The first and the second terms are used to achieve vertical and horizontal attention consistency, respectively. Take ${\rm\textbf{M}}^{s}_{\emph{right}\rightarrow\emph{left}}$ as an example, ${\rm\textbf{M}}^{s}_{\emph{right}\rightarrow\emph{left}}$$(i,j,k)$ measures the contribution of $(i,k)$ in ${\rm\textbf{I}}^{s}_\emph{right}$ to $(i,j)$ in ${\rm\textbf{I}}^{s}_\emph{left}$ using their feature similarity  $Sim(\textbf{F}^{s}_\emph{left}(i,j),\textbf{F}^{s}_\emph{right}(i,k))$. Our smoothness loss enforces $Sim(\textbf{F}^{s}_\emph{left}(i\!+\!1,j),\textbf{F}^{s}_\emph{right}(i\!+\!1,k))$ and $Sim(\textbf{F}^{s}_\emph{left}(i,j\!+\!1),\textbf{F}^{s}_\emph{right}(i,k\!+\!1))$ to be close to $Sim(\textbf{F}^{s}_\emph{left}(i,j),\textbf{F}^{s}_\emph{right}(i,k))$. Consequently, smoothness in correspondence (disparity) space can be encouraged. 
	
	\noindent$\bullet$ {$\mathcal{L}^{s}_{\rm{PAM\emph{-}c}}$}
	\vspace{1.5mm}

	In addition to $\mathcal{L}^{s}_{\rm{PAM\emph{-}p}}$ and $\mathcal{L}^{s}_{\rm{PAM\emph{-}s}}$, we further introduce a cycle loss to achieve cycle consistency. Since ${\rm\textbf{M}}^{s}_{\emph{left}\rightarrow\emph{right}\rightarrow\emph{left}}$ and ${\rm\textbf{M}}^{s}_{\emph{right}\rightarrow\emph{left}\rightarrow\emph{right}}$ in Eq. (\ref{eq3}) can be considered as identity matrices, we design a cycle loss as:
	\begin{equation}
	\begin{aligned}
	\mathcal{L}^{s}_{\rm{PAM\emph{-}c}}=&\frac{1}{N^{s}_\emph{left}}\sum_{p\in{{\rm\textbf{V}}^{s}_\emph{left}}}\left\Vert {\rm\textbf{M}}^{s}_{{\emph{left}\rightarrow\emph{right}\rightarrow\emph{left}}}(p)-I^{s}(p)\right\Vert_{1}+\\
	&\frac{1}{N^{s}_\emph{right}}\sum_{p\in{{\rm\textbf{V}}^{s}_\emph{right}}}\left\Vert{\rm\textbf{M}}^{s}_{{\emph{right}\rightarrow\emph{left}\rightarrow\emph{right}}}(p)-I^{s}(p)\right\Vert_{1},
	\end{aligned}
	\end{equation}
	where  $I^{s}$ is a stack of identity matrices.
	
	In summary, the overall loss is defined as:
	\begin{equation}
	\begin{aligned}
	\mathcal{L}_{\rm{unsup}}=&\mathcal{L}_{\rm{p}}+\lambda_{\rm{s}}\mathcal{L}_{\rm{s}}+\\
	&\lambda_{\rm{PAM}}(0.2\mathcal{L}^1_{\rm{PAM}}+0.3\mathcal{L}^{2}_{\rm{PAM}}+0.5\mathcal{L}^3_{\rm{PAM}}).
	\end{aligned}
	\end{equation}
	Note that, groundtruth disparities are not used in the overall loss. That is, our network is trained in an unsupervised manner.

	\subsection{Experimental Results}
	\subsubsection{Datasets and Metrics}
	We used the SceneFlow \cite{Mayer2016Large} and KITTI 2015 \cite{Menze2015Object} datasets for training and test. 
	
	\textbf{SceneFlow}: The SceneFlow dataset is a synthetic dataset consisting of 35454 training image pairs and 4370 test image pairs of size \textcolor{black}{$540\times960$}.
	
	\textbf{KITTI 2015}: The KITTI 2015 dataset is a real-world dataset with street views collected from a driving car. This dataset contains 200 training image pairs and 200 test image pairs of size \textcolor{black}{$375\times1242$}.

	For evaluation, we used end-point-error (EPE) and t-pixel error rate ($>\!{\rm t}px$) as metrics. Metrics for both non-occluded regions (Noc) and all pixels (All) are evaluated. 
	
	\subsubsection{Implementation Details}
	We first trained our network on the SceneFlow dataset and performed ablation study on this dataset. During training phase, patches of size $256\times512$ were randomly cropped as inputs. $\lambda_{\rm{PAM\emph{-}s}}$, $\lambda_{\rm{PAM\emph{-}c}}$, $\lambda_{\rm{s}}$ and $\lambda_{\rm{PAM}}$ are set to 1, 1, 0.1, and 1, respectively. All models were optimized using the Adam method \cite{Kingma2015Adam} with $\beta_{1}=0.9$, $\beta_{2}=0.999$ and a batch size of 14. The initial learning rate was set to $1\!\times\!10^{-3}$ for 5 epochs and decreased to $1\!\times\!10^{-4}$ for another 5 epochs. Note that, pixels with disparities over 192 are excluded for training to demonstrate the generalization of our network to large disparity variations, since our test set contains image pairs with disparities over 192. Following \cite{Li2018Occlusion}, pixels with disparities over 192 are also excluded for evaluation unless specified otherwise. Next, we fine-tuned our network on the KITTI 2015 dataset to obtain the final model for submission.  $\lambda_{\rm{PAM\emph{-}s}}$, $\lambda_{\rm{PAM\emph{-}c}}$, $\lambda_{\rm{s}}$, and $\lambda_{\rm{PAM}}$ are set to 5, 5, 0.5, and 1 \textcolor{black}{for} this dataset, respectively. The initial learning rate was set to $1\!\times\!10^{-4}$ for 60 epochs and decreased to $1\!\times\!10^{-5}$ for another 20 epochs. All experiments were conducted on a PC with two Nvidia RTX 2080Ti GPUs.
	
	\begin{table}[t]
		\caption{Comparative results achieved on SceneFlow by our PASMnet trained with different settings.}
		\label{tab1-1}
		\begin{center}
			\footnotesize
			\setlength{\tabcolsep}{0.8mm}{
				\begin{tabular}{|c|ccc|c|c|c|}
					\hline 
					Model & Hourglass  & Coarse-to-Fine & \# Blocks & EPE & $>1px$ & $>3px$
					\tabularnewline
					\hline
					PASMnet &            & \checkmark & 1 & 4.86 & 19.58 & 16.31 \tabularnewline 
					PASMnet & \checkmark &            & 1 & 4.78 & 19.50 & 16.21 \tabularnewline 
					PASMnet & \checkmark & \checkmark & 1 & 4.68 & 19.32 & 16.09 \tabularnewline 
					PASMnet & \checkmark & \checkmark & 2 & 4.63 & 19.19 & 16.05 \tabularnewline 
					PASMnet & \checkmark & \checkmark & 3 & 4.58 & 19.05 & 15.97 \tabularnewline 
					PASMnet & \checkmark & \checkmark & 4 & \textbf{4.54} & \textbf{18.99} & \textbf{15.91} \tabularnewline	
					\hline	
			\end{tabular}}
		\end{center}
	\end{table}

	\subsubsection{Ablation Study}
	
	\noindent \textbf{Hourglass Feature Extraction.}
	Hourglass network is used to aggregate features from multiple scales in our PASMnet. To demonstrate its effectiveness, we replaced the hourglass network with a pyramid network for feature extraction. It can be observed from Table~\ref{tab1-1} that the 1-pixel/3-pixel error rates are increased from 19.32/16.09 to 19.58/16.31 if hourglass network is replaced with a pyramid network. As compared to pyramid network, hourglass network can aggregate features from multiple scales, resulting in features rich of multi-scale geometric information. Therefore, better performance can be achieved.
	
	\noindent \textbf{Coarse-to-fine Manner.}
	Our cascaded parallax-attention module regresses matching costs in a coarse-to-fine manner. To demonstrate its effectiveness, we introduced a variant that regresses matching costs on a single scale (\emph{i.e.}, $\times\frac{1}{4}$). It can be observed from Table~\ref{tab1-1} that if matching costs are regressed on a single scale, the 1-pixel/3-pixel error rates are increased from 19.32/16.09 to 19.50/16.21. By using a coarse-to-fine manner, disparities can be progressively refined and better performance can be achieved.
		
	\begin{table}[t]
		\caption{\textcolor{black}{Ablation results for the disparity refinement module in our PASMnet on SceneFlow.}}
		\label{tab1-10}
		\begin{center}
			\footnotesize
			\setlength{\tabcolsep}{3mm}{
				\begin{tabular}{|l|c|c|c|}
					\hline 
					Model & EPE & $>1px$ & $>3px$
					\tabularnewline				
					\hline
					PASMnet\_wo\_Refinement  & 4.56 & 19.15 & 16.11	
					\tabularnewline
					\hline
					PASMnet 	& \textbf{4.54} & \textbf{18.99} & \textbf{15.91}		
					\tabularnewline
					\hline				
			\end{tabular}}
		\end{center}
	\end{table}
	
\begin{figure}[t]
		\centering
		\includegraphics[width=1\linewidth]{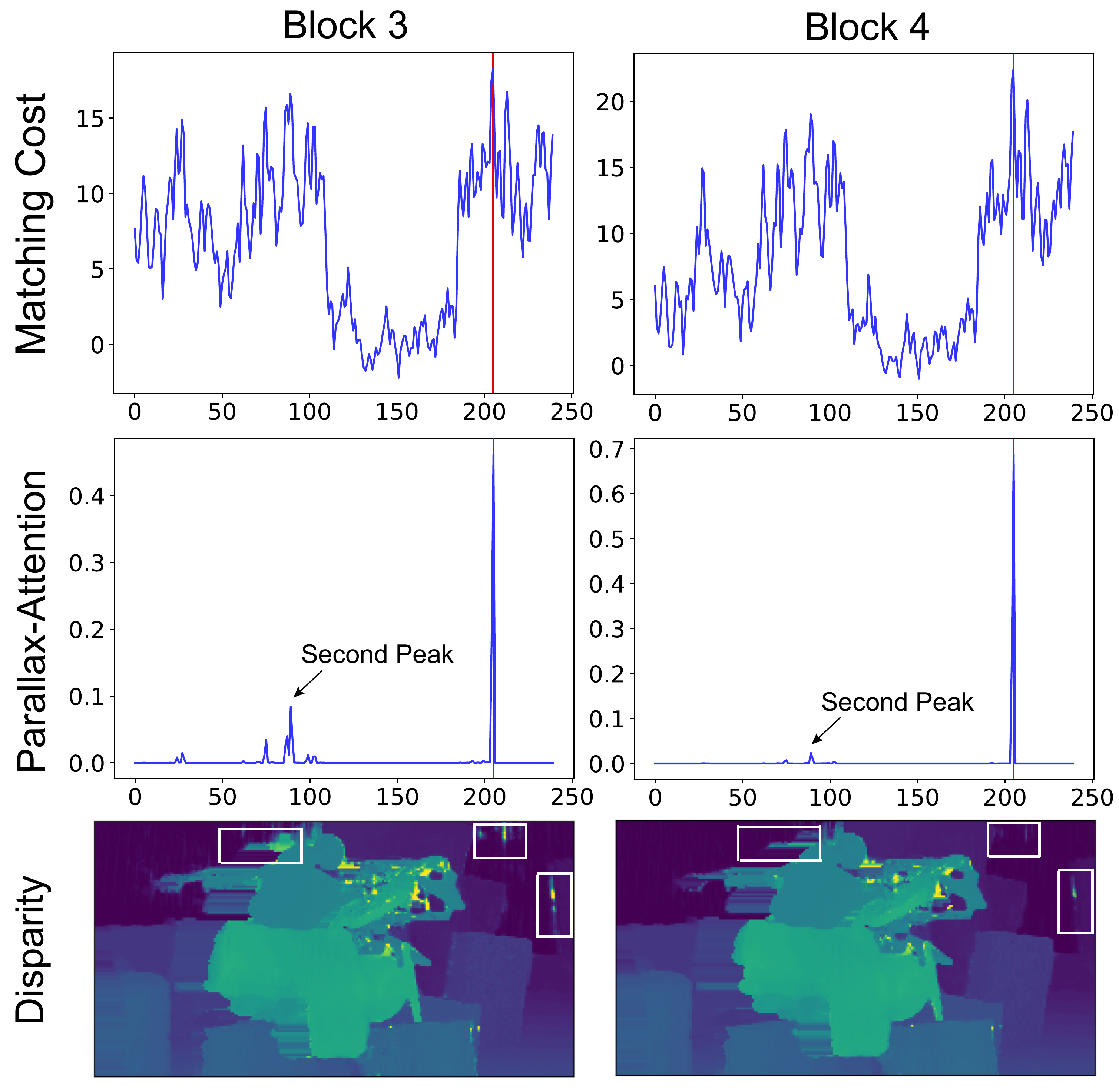}
		\caption{\textcolor{black}{Matching cost distributions, parallax-attention distributions and disparities produced by blocks 3 and 4 in stage 3 of our cascaded parallax-attention module. The red lines in the first and second rows represent the groundtruth disparity. As compared to block 3, block 4 produces more accurate and smoother disparities by cascading more parallax-attention blocks, especially in regions denoted with white boxes.}}
		\label{fig15}
	\end{figure}

	\noindent \textbf{Number of Cascaded Blocks.}
	We tested the performance of our PASMnet with different number of parallax-attention blocks. From Table~\ref{tab1-1} we can see that, the performance of our network improves as the number of parallax-attention blocks increases. Specifically, our PASMnet with 4 blocks outperforms the variant with 1 block, with 1-pixel/3-pixel error rates being improved from 19.32/16.09 to 18.99/15.91. That is because, by cascading more blocks, our network can progressively refine the matching cost for better performance.
	\textcolor{black}{We further fed the matching costs produced by the last two parallax-attention blocks in stage 3 (\emph{i.e.}, blocks 3 and 4) to the output module for disparity regression. The disparities, matching cost distributions and parallax-attention distributions are visualized in Fig.~\ref{fig15}. As more parallax-attention blocks are cascaded, matching cost is aggregated to produce higher peak values at the groundtruth disparity in the parallax-attention distributions. Therefore, more accurate and smoother disparities are produced.}
	
\begin{figure}[t]
		\centering
		\includegraphics[width=1\linewidth]{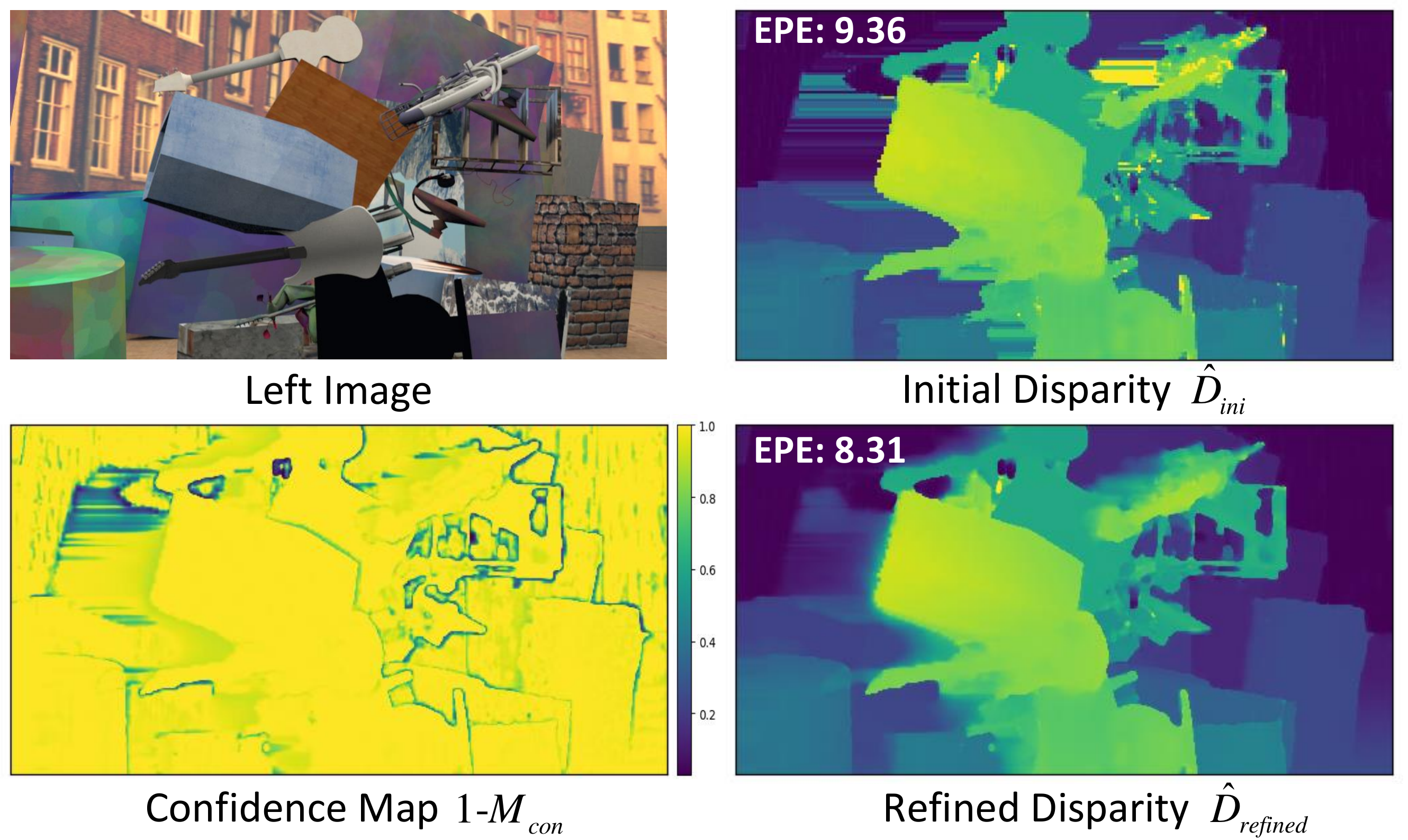}
		\caption{\textcolor{black}{An example of initial disparity $\rm{\hat{\textbf{D}}}_\emph{ini}$, confidence map $1\!-\!M_\emph{con}$ and refined disparity $\rm{\hat{\textbf{D}}}_\emph{refined}$ produced by our PASMnet.}}
		\label{fig16}
	\end{figure}

	\noindent \textbf{\textcolor{black}{Disparity Refinement.}}
	\textcolor{black}{
		After the cascaded parallax-attention module, disparity refinement is performed by using features from the left image to provide structural information. To demonstrate the effectiveness of disparity refinement, we introduced a variant by removing this disparity refinement module and upsampling ${\rm\hat{\textbf{D}}}_\emph{ini}$ to obtain the final disparity. It can be observed from Table~\ref{tab1-10} that the 1-pixel/3-pixel error rates are increased from 18.99/15.91 to 19.15/16.11 without disparity refinement. That is because, structural information helps to effectively refine the disparity. We further visualize the confidence map learned within the disparity refinement module in Fig.~\ref{fig16}. We can see that occluded regions and edge regions benefit a lot from disparity refinement while other regions in the initial disparity $\rm{\hat{\textbf{D}}}_\emph{ini}$ are ``good enough''.  
	}

	\noindent\textcolor{black}{\textbf{PAM vs. Cost Volume.}}
	To demonstrate the efficiency of the proposed PAM, 
	\textcolor{black}{we compare our parallax-attention block with 3D convolution in terms of model size, FLOPs and memory consumption in Table~\ref{tab1-11}. With the same number of channels, our parallax-attention block has fewer parameters than 3D convolution ($20C$ vs. $27C$). Assume that $D\!=\!192,H\!=\!540,W\!=\!960$ (\emph{i.e.}, the size of images in SceneFlow), when performed at the $\frac{1}{4}$ resolution level (\emph{i.e.}, $D\!=\!48,H\!=\!135,W\!=\!240$), our block achieves a $16.2\times$ reduction in FLOPs and a comparable memory cost as compared to 3D convolution when $C\!=\!12$. As $C$ increases, better efficiency is achieved by our parallax-attention block, with $23.5\times$/$2.1\times$ reductions in FLOPs/memory cost being achieved for $C=32$.}
	
	\textcolor{black}{We further introduced a network variant by replacing PAM with a 4D cost volume technique and then compared the stereo matching performance of this variant to our PASMnet.} Specifically, we replaced the four cascaded parallax-attention blocks at each stage of our cascaded parallax-attention module with four 3D convolutions. \textcolor{black}{Note that, the number of channels are adjusted to ensure that this variant has comparable parameters to our PASMnet.} The maximum disparity is set to 192 for the cost volume technique. Comparative results achieved on SceneFlow are shown in Table \ref{tab1-2}. Compared to the 4D cost volume technique, our PAM achieves much better performance with comparable model size and inference time. \textcolor{black}{With the proposed PAM loss,} we can directly regularize our PAM to capture accurate and consistent stereo correspondence. Therefore, better performance can be achieved. Moreover, our PAM also achieves higher efficiency, with its memory consumption being half of the cost volume technique.

	\begin{table}[t]
		\caption{\textcolor{black}{Comparison between our parallax-attention block and 3D convolution. Here, $\otimes$ refers to geometry-aware matrix multiplication.}}
		\label{tab1-11}
		\begin{center}
			\footnotesize
			\setlength{\tabcolsep}{0.3mm}{
				\begin{tabular}{|ll|l|l|l|}
					\hline 
					\multicolumn{2}{|l|}{Module} & Params. & FLOPs & Memory
					\tabularnewline
					\hline
					\multirow{4}{*}{\tabincell{l}{Parallax-\\Attention\\Block}}
					&\vline\,$3\!\times\!3$ Conv
					&$C^2\!\times\!3^2\!\times\!2$
					&$HWC^2\!\times\!3^2\!\times\!2\!\times\!2$
					&$HW\!\times\!C\!\times2\!\times\!2$
					\tabularnewline
					&\vline\,$1\!\times\!1$ Conv 
					&$C^2\!\times\!2$
					&$HWC^2\!\times\!2\!\times\!2$
					&$HW\!\times\!C\!\times2\!\times2$
					\tabularnewline
					&\vline\,$\otimes$&-
					&$HWC\!\times\!W\!\times2$
					&$HW\!\times\!W\!\times\!2$	
					\tabularnewline
					&\vline\,Total&$C^2\!\times\!20$
					&$HWC^2\!\times\!(40\!+\!\frac{2W}{C})$	
					&$HW\!\times\!(8C\!+\!2W)$	
					\tabularnewline
					\hline
					3D Conv
					&
					&$C^2\!\times\!3^3$
					&$HWC^2\!\times\!3^3\!\times\!D$
					&$HW\!\times\!C\!\times\!D$	
					\tabularnewline
					\hline
			\end{tabular}}
		\end{center}
	\end{table}

	\begin{table}[t]
		\caption{Comparison between our PAM and cost volume formation on SceneFlow.}
		\label{tab1-2}
		\begin{center}
			\footnotesize
			\setlength{\tabcolsep}{1mm}{
				\begin{tabular}{|l|c|c|c|c|c|c|}
					\hline 
					Model & Params. & Time & Memory & EPE & $>1px$ & $>3px$
					\tabularnewline
					\hline
					Cost Volume  & 7.35M & $1\times$ & $1\times$ & 6.02 & 19.79 &	17.42	
					\tabularnewline
					\hline
					PAM 		 & 7.61M & $1.1\!\times$   & $0.5\!\times$   & \textbf{4.54} & \textbf{18.99} & \textbf{15.91}		
					\tabularnewline
					\hline
			\end{tabular}}
		\end{center}
	\end{table}

	\begin{table}[!t]
		\caption{Comparative results achieved on SceneFlow by our PASMnet trained with different losses.}
		\label{tab1-3}
		\begin{center}
			\footnotesize
			\setlength{\tabcolsep}{0.8mm}{
				\begin{tabular}{|c|ccccc|c|c|c|}
					\hline 
					Model & $\mathcal{L}_{\rm{p}}$ & $\mathcal{L}_{\rm{s}}$ & $\mathcal{L}_{\rm{PAM-p}}$ & $\mathcal{L}_{\rm{PAM-s}}$ & $\mathcal{L}_{\rm{PAM-c}}$ & EPE & $>1px$ & $>3px$
					\tabularnewline
					\hline
					PASMnet & \checkmark &            &            &            &            & 6.07 & 20.78 & 17.11  \tabularnewline 
					PASMnet & \checkmark & \checkmark &            &            &            & 4.87 & 19.69 & 16.54  \tabularnewline 
					PASMnet & \checkmark & \checkmark & \checkmark &            &            & 4.63 & 19.46 & 16.12  \tabularnewline 
					PASMnet & \checkmark & \checkmark & \checkmark & \checkmark &            & 4.58 & 19.13 & 15.99  \tabularnewline 
					PASMnet & \checkmark & \checkmark & \checkmark & \checkmark & \checkmark & \textbf{4.54} & \textbf{18.99} & \textbf{15.91}  \tabularnewline 
					\hline
			\end{tabular}}
		\end{center}
		\vspace{-0.145cm}
	\end{table}

	\begin{figure}[t]
		\centering
		\includegraphics[width=1\linewidth]{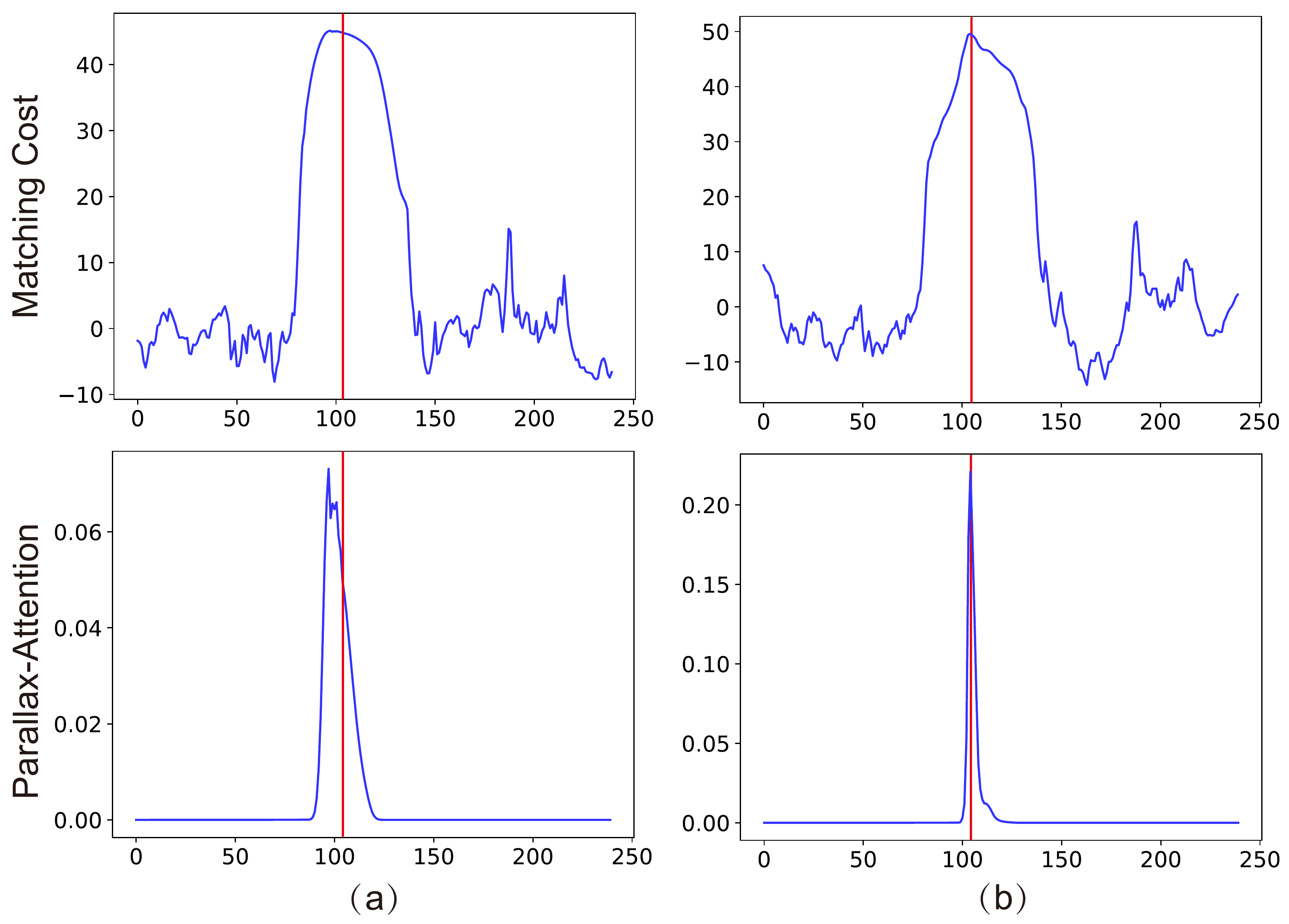}
		\caption{An example of matching cost and parallax-attention distributions. (a) Results produced by our PASMnet trained without $\mathcal{L}_{\rm{PAM}}$. (b) Results produced by our PASMnet. The red lines represent the groundtruth disparity. Using $\mathcal{L}_{\rm{PAM}}$ for regularization, our PASMnet produces more reasonable cost and parallax-attention distributions, which have peak values at the groundtruth disparity with higher sharpness.}
		\label{fig13}
	\end{figure}

	\begin{table}[t]
		\caption{Comparison between PAM and cost volume formation on SceneFlow with different resolutions.}
		\begin{center}
			\footnotesize
			\setlength{\tabcolsep}{1.2mm}{
				\begin{tabular}{|l|ccc|ccc|ccc|}
					\hline 
					\multirow{2}{*}{Resolution} & \multicolumn{3}{c|}{Cost Volume} & \multicolumn{3}{c|}{PAM}
					\tabularnewline
					& EPE & $>1px$ & $>3px$ & EPE  & $>1px$ & $>3px$
					\tabularnewline
					\hline
					$960\times540$  & 44.04 & 48.53 & 47.04 & \textbf{22.02} & \textbf{33.17} & \textbf{31.43}
					\tabularnewline
					$480\times270$	& 14.60 & 35.75 & 31.77 & \textbf{10.57} & \textbf{35.56} & \textbf{30.85}
					\tabularnewline				
					$240\times135$ 	& 7.87  & \textbf{39.00} & 30.31 & \textbf{6.04} & 39.93 & \textbf{29.64}
					\tabularnewline		
					\hline	
					Average     	& 22.17 & 41.09 & 36.37 & \textbf{12.88} & \textbf{36.22} & \textbf{30.64}
					\tabularnewline
					\hline
			\end{tabular}}
		\end{center}
		\label{tab1-4}
	\end{table}

	\noindent \textbf{Losses.}
	We retrained our PASMnet using different losses to test the effectiveness of our loss function. It can be observed from Table~\ref{tab1-3} that the EPE/1-pixel error rate/3-pixel error rate achieved by our network are increased from 4.54/18.99/15.91 to 6.07/20.78/17.11 if only photometric loss is used for training. That is because, photometric loss cannot handle texture-less regions well. If smoothness loss is included for training, the EPE value is decreased to 4.87. Further, the performance is gradually improved if PAM loss is added. That is because, PAM loss regularizes our PAM to capture accurate and consistent stereo correspondence.
	
	An example of matching cost and parallax-attention distributions produced by our network trained with different losses is shown in Fig.~\ref{fig13}. Using $\mathcal{L}_{\rm{PAM}}$ for regularization, our network produces more reasonable matching cost and parallax-attention distributions. Both of them have peak values at the groundtruth disparity with high sharpness. This clearly demonstrates that performing direct regularization on parallax-attention maps enables our PAM to achieve better performance.

	\subsubsection{\textcolor{black}{Flexibility to Disparity Variations}}
	We tested the flexibility of our PAM with respect to large disparity variations. We chose 20 image pairs from the test set of SceneFlow with more than 20\% disparity values larger than 200 for evaluation. Note that, all pixels were used for evaluation.
	
	\noindent \textbf{Resolutions.} We resized chosen image pairs to different resolutions to test the flexibility of our PAM. Results achieved on images with different resolutions are shown in Table \ref{tab1-4}. It can be observed that our PAM outperforms the cost volume technique on all metrics except 1-pixel error rate at the resolution of $240\times135$. Moreover, the performance improvement achieved by our PAM is enhanced as the resolution increases. That is because, the fixed maximum disparity hinders the cost volume technique to capture longer-range correspondence. In contrast, our PAM is more flexible and robust to large disparity variations and achieves better performance.
	\begin{figure*}[t]
		\centering
		\includegraphics[width=1\linewidth]{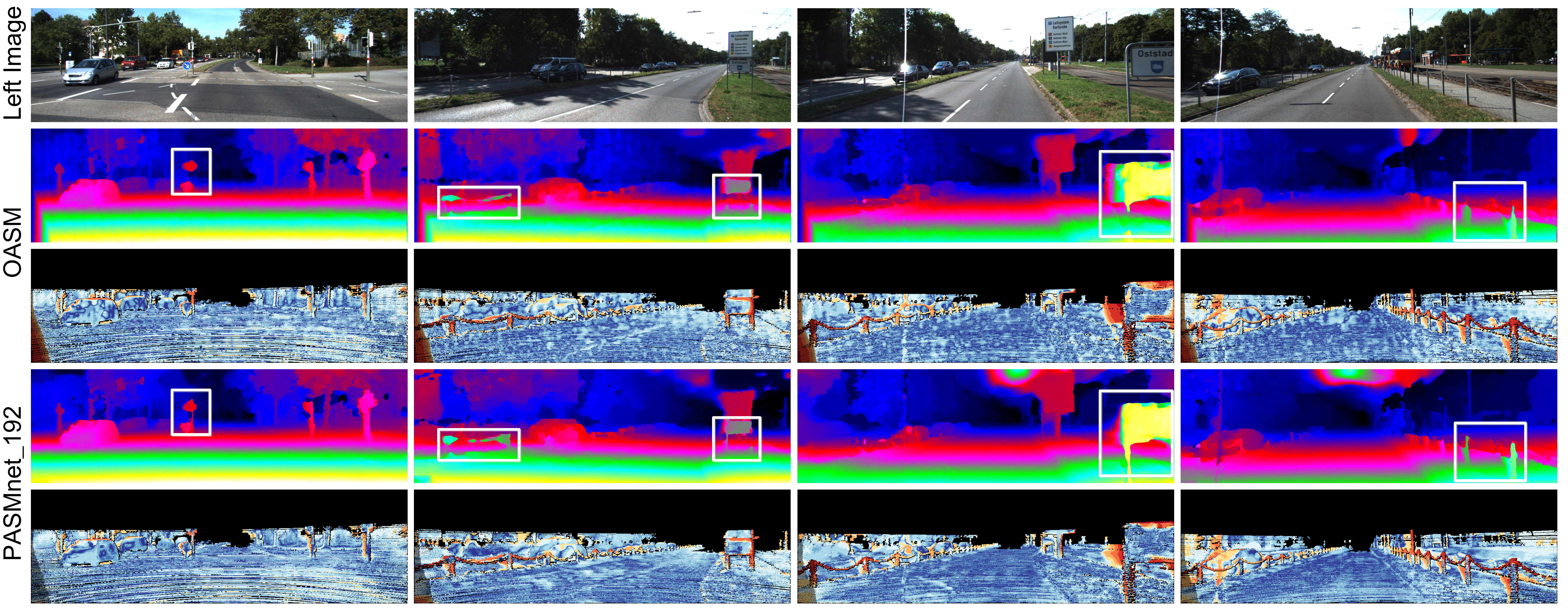}
		\caption{Results achieved on the KITTI 2015 dataset.}
		\label{fig12}
	\end{figure*}
	
	\begin{table}[t]
		\caption{Comparison between PAM and cost volume formation on SceneFlow with different depths.}
		\begin{center}
			\footnotesize
			\setlength{\tabcolsep}{1.2mm}{
				\begin{tabular}{|l|ccc|ccc|ccc|}
					\hline 
					\multirow{2}{*}{Disparity} & \multicolumn{3}{c|}{Cost Volume} & \multicolumn{3}{c|}{PAM}
					\tabularnewline
					& EPE & $>1px$ & $>3px$ & EPE  & $>1px$ & $>3px$
					\tabularnewline
					\hline
					$>$200  & 150.21 & 99.99 & 99.99 & \textbf{57.90} & \textbf{41.62} & \textbf{41.60}
					\tabularnewline
					100-200	& 36.24 & 45.80 & 45.79 & \textbf{25.16} & \textbf{33.15} & \textbf{33.15}
					\tabularnewline				
					$<$100 	& \textbf{13.71} & \textbf{33.92} & 31.92 & 16.31 & 34.57 & \textbf{31.81}
					\tabularnewline
					\hline
					Average & 44.04 & 48.53 & 47.04 & \textbf{22.02} & \textbf{33.17} & \textbf{31.43}
					\tabularnewline
					\hline
			\end{tabular}}
		\end{center}
		\label{tab1-5}
	\end{table}	
	\noindent \textbf{Depths.} We compared the performance on regions with different depths (disparities) to test the flexibility of our PAM. Results achieved on images with different depths are shown in Table \ref{tab1-5}. It can be observed that the performance of our PAM is significantly better than the cost volume technique on regions with disparities larger than 100. That is because, our PAM can get rid of setting a fixed maximum disparity to handle large disparities. In contrast, the cost volume technique cannot capture the correspondences with disparities larger than the fixed maximum disparity. 

	\begin{table}[t]
		\caption{\textcolor{black}{Comparison to existing supervised and unsupervised stereo matching methods on KITTI 2015.}}
		\label{tab1-6}
		\begin{center}
			\footnotesize
			\setlength{\tabcolsep}{0.9mm}{
				\begin{tabular}{|c|l|ccc|ccc|}
					\hline 
					& & \multicolumn{3}{|c|}{Noc} & \multicolumn{3}{|c|}{All} \tabularnewline
					& & D1-bg & D1-fg & D1-all & D1-bg & D1-fg & D1-all \tabularnewline
					\hline
					\multirow{6}{*}{\rotatebox{90}{Supervised}} 
					& DispNet \cite{Mayer2016Large}     		&4.11 &3.72  &4.05 &4.32 &4.41  &4.34 	\tabularnewline
					& GC-Net  \cite{Kendall2017End}       		&2.02 &5.58  &2.61 &2.21 &6.16  &2.87	\tabularnewline
					& CRL     \cite{Pang2017Cascade} 			&2.32 &3.12  &2.45 &2.48 &3.59  &2.67	\tabularnewline
					& iResNet \cite{Liang2018Learning}          &2.07 &\textbf{2.76}  &2.19 &2.25 &\textbf{3.40}  &2.44	\tabularnewline
					& PSMNet  \cite{Chang2018Pyramid}           &\textbf{1.71} &4.31  &\textbf{2.14} &\textbf{1.86} &4.62  &\textbf{2.32}	\tabularnewline
					\cline{2-8}
					& PASMnet\_192 (ours)        &1.88&3.91&2.22&2.04&4.33&2.41	\tabularnewline
					\hline
					\multirow{7}{*}{\rotatebox{90}{Unsupervised}} 
					& USCNN       	\cite{Ahmadi2016Unsupervised}    &-    &-     &11.71&-    &-     &16.55	\tabularnewline
					& Yu et al.  	\cite{Yu2016Back}                &-    &-     &8.35 &-    &-     &19.14	\tabularnewline
					& Zhou et al.	\cite{Zhou2017Unsupervised}      &-    &-     &8.61 &-    &-     &9.91	\tabularnewline
					& SegStereo 	\cite{Yang2018Segstereo} 		 &-    &-     &7.70 &-    &-     &8.79	\tabularnewline
					& OASM 			\cite{Li2018Occlusion}           &5.44 &17.30 &7.39 &6.89 &19.42 &8.98	\tabularnewline
					\cline{2-8}
					& PASMnet (ours)             & 5.35 & 15.24 & 6.99 & 5.89 & 16.74 & 7.70	\tabularnewline
					& PASMnet\_192 (ours)        &\textbf{5.02} &\textbf{15.16} &\textbf{6.69} &\textbf{5.41} &\textbf{16.36} &\textbf{7.23}	\tabularnewline
					\hline
			\end{tabular}}
		\end{center}
	\end{table}
	
	\subsubsection{Comparison to State-of-the-arts}
	
	We compare our PASMnet with existing unsupervised stereo matching methods \cite{Ahmadi2016Unsupervised,Yu2016Back,Zhou2017Unsupervised,Yang2018Segstereo,Li2018Occlusion} on the KITTI 2015 dataset. Note that, \cite{Yang2018Segstereo} and \cite{Li2018Occlusion} use a maximum disparity of 192 as a prior knowledge for both training and test. For fair comparison, we also trained a variant with a maximum disparity of 192. Specifically, we excluded positions with disparities larger than the maximum disparity in the attention maps. \textcolor{black}{Moreover, we include several supervised stereo matching methods \cite{Mayer2016Large,Kendall2017End,Pang2017Cascade,Liang2018Learning,Chang2018Pyramid} \textcolor{black}{and trained our PASMnet\_192 in a supervised manner} for comparison.} \textcolor{black}{Specifically, we first finetuned our PASMnet\_192 (which was trained on SceneFlow in an unsupervised manner) on SceneFlow using groundtruth disparities as its supervision. Then, this model was further finetuned on the KITTI 2015 dataset.} As we can see from Table~\ref{tab1-6}, \textcolor{black}{our PASMnet outperforms other unsupervised methods by notable margins and significantly reduces the performance gap between supervised and unsupervised methods. With a prior maximum disparity, the performance of our PASMnet is further improved.} Specifically, our PASMnet\_192 achieves a much lower 3-pixel error rate (6.69/7.23) than OASM (7.39/8.98). \textcolor{black}{If supervision is provided, our PASMnet\_192 produces competitive results to iResNet and PSMNet.} Figure~\ref{fig12} shows several visual examples. From the white boxes in Fig.~\ref{fig12}, it can be observed that our PASMnet produces much \textcolor{black}{more accurate} and smoother results.

	\begin{figure*}[ht]
		\centering
		\includegraphics[width=1\linewidth]{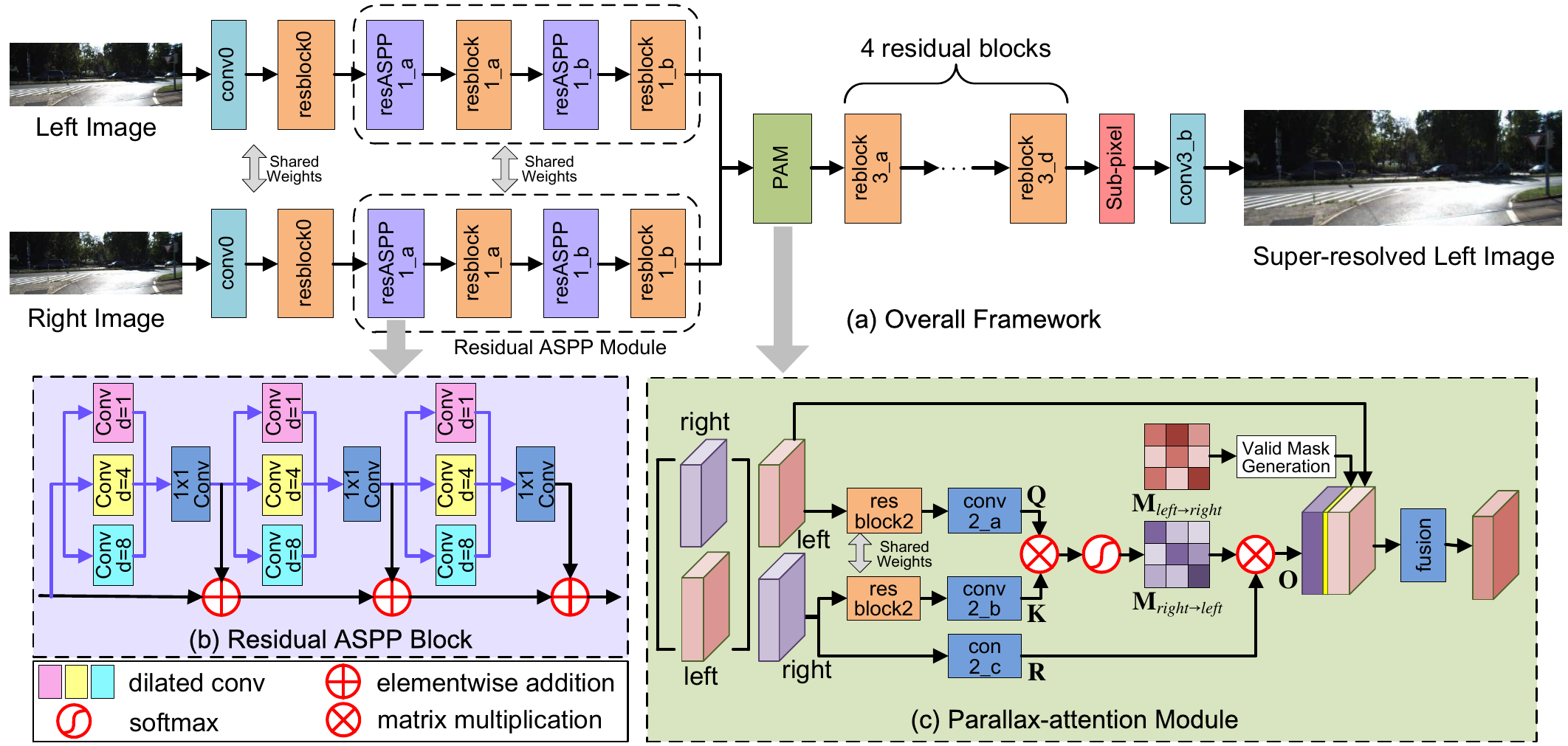}
		\caption{An overview of our PASSRnet.}
		\label{fig6}
	\end{figure*}
	
	\section{PAM for Stereo Image Super-Resolution}
	\label{Sec5}
	Stereo image SR aims at recovering an HR left image from a pair of LR stereo images. The key challenge for stereo image SR lies in exploiting stereo correspondence to aggregate information from the pair of stereo images. Our PAM can effectively aggregate information from a pair of stereo images using the parallax-attention map as its guidance. Therefore, the PAM can be applied for the stereo image SR task.
	
	\subsection{Network Architecture}
	\subsubsection{Overview}

	Based on our PAM, we propose a parallax-attention stereo super-resolution network (PASSRnet) for stereo image SR. Specifically, the stereo images are first fed to a convolutional layer and a residual block for feature extraction. Then, the resulting features are passed to a residual atrous spatial pyramid pooling (ASPP) module to extract deep feature representations. Next, a parallax-attention module is used to aggregate features from a stereo pair. Finally, 4 residual blocks and a sup-pixel convolutional layer are used to generate the SR results. The architecture of our PASSRnet is shown in Fig. \ref{fig6}.

	\subsubsection{Residual Atrous Spatial Pyramid Pooling (ASPP) Module}
	
	Since feature representation with rich context information is beneficial to correspondence estimation \cite{Chang2018Pyramid}, we propose a residual ASPP module to enlarge the receptive field and extract hierarchical features with dense pixel sampling rate and scales. As shown in Fig. \ref{fig6} (a), input features are first fed to a residual ASPP block to generate multi-scale features. The resulting features are then sent to a residual block for feature fusion. This structure is repeated twice to produce final features. Within each residual ASPP block (as shown in Fig. \ref{fig6} (b)), we first combine three dilated convolutions (with dilation rates of 1, 4, 8) to form an ASPP group, and then cascade three ASPP groups in a residual manner. Our residual ASPP block not only enlarges the receptive field, but also enriches the diversity of convolutions, resulting in an ensemble of convolutions with different receptive regions and dilation rates. The highly discriminative feature learned by our residual ASPP module is beneficial to the overall SR performance, as demonstrated in Sec.~\ref{Sec5.4}.
	
	\subsubsection{Parallax-Attention Module}
	
	Based on our PAM, we introduced a parallax-attention module to exploit stereo correspondence for the aggregation of features from a pair of stereo images. As shown in Fig.~\ref{fig6}(c), features from left/right images are first fed to a transition residual block for feature adaption. Then, the resulting features are used to generate the output feature $\textbf{O}$ and the parallax-attention map $\rm\textbf{M}_{\emph{right}\rightarrow\emph{left}}$ based on PAM. Next, features from left/right images are exchanged to produce $\rm\textbf{M}_{\emph{left}\rightarrow\emph{right}}$ for valid mask generation. Finally, the concatenation of the output feature, the feature from the left image and the valid mask is fed to a $1\times1$ convolution for feature fusion. 
	
	It should be noted that our PASSRnet can be considered as a multi-task network to learn both stereo correspondence and SR. However, using shared features for different tasks usually suffers from training conflict \cite{Sener2018Multi}. Therefore, a transition block is used in our parallax-attention module to alleviate this problem. The effectiveness of the transition block is demonstrated in Sec.~\ref{Sec5.4}.
	
	\subsection{Losses}
	We used the mean square error (MSE) loss as the SR loss:
	\begin{equation}
	\mathcal{L_{\mathrm{SR}}}=\left\Vert{\rm\textbf{I}}_\emph{left}^{SR}-{\rm\textbf{I}}_\emph{left}^{H}\right\Vert _{2}^{2},
	\end{equation}
	where ${\rm\textbf{I}}_\emph{left}^{SR}$ and ${\rm\textbf{I}}_\emph{left}^{H}$ represent the SR result and HR groundtruth of the left image, respectively. Other than the SR loss, we also used the PAM loss (as defined in Sec.~\ref{Sec4.2.3}) to help our network exploit the correspondence between stereo images. The overall loss is formulated as:
	\begin{equation}
	\mathcal{L}=\mathcal{L}_{\mathrm{SR}}+\lambda\mathcal{L}_{\rm{PAM}},
	\end{equation}
	where $\lambda$ is empirically set to 0.005. The performance of our network with different losses will be analyzed in Sec. \ref{Sec5.4}. Note that, stereo correspondence is learned in an unsupervised manner using $\mathcal{L}_{\rm{PAM}}$ while SR is learned in a supervised manner using $\mathcal{L}_{\rm{SR}}$.

	\subsection{The Flickr1024 Dataset}
\begin{figure}[tp]
		\centering
		\includegraphics[width=0.95\linewidth]{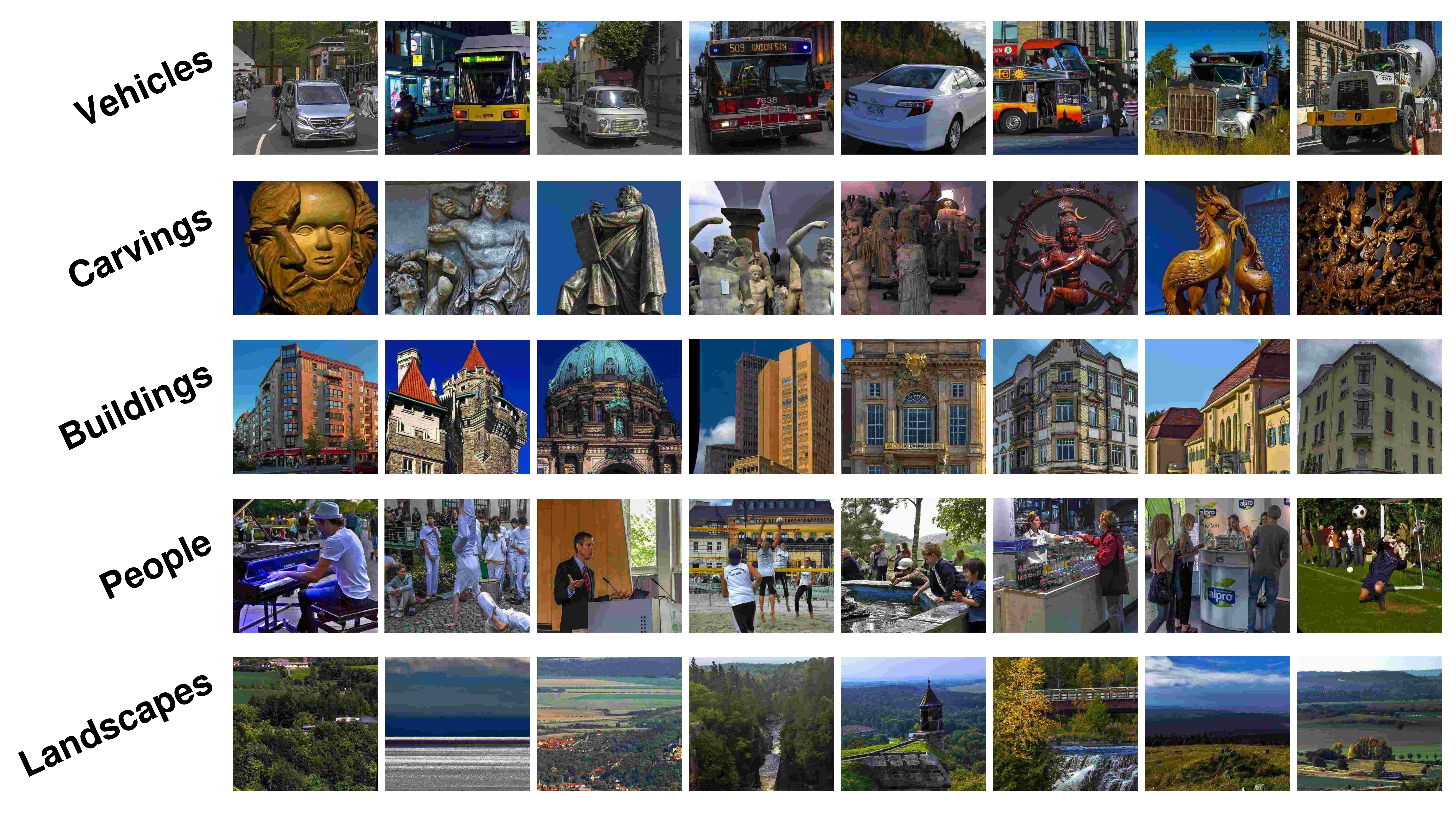}
		\caption{Samples of different scenes covered in the Flickr1024 dataset.}
		\label{fig8}
	\end{figure}

	Although several stereo datasets such as Middlebury \cite{Scharstein2014High} and KITTI \cite{Geiger2012Are,Menze2015Object} are already available, these datasets are mainly proposed for stereo matching. Further, the Middlebury dataset only consists of close shots of man-made objects, while the KITTI 2012 and KITTI 2015 datasets only consist of road scenes. For stereo image SR task, a large dataset which covers diverse scenes and consists of images with high quality and rich details is required. Therefore, we introduce a new Flickr1024 dataset \cite{Wang2019Flickr1024} for stereo image SR, which covers a large diversity of scenes, including landscapes, urban scenes, people and man-made objects, as shown in Fig. \ref{fig8}. Our Flickr1024 dataset is available at: \emph{\color{blue}{https://yingqianwang.github.io/Flickr1024}}.
	
	\subsubsection{Data Collection and Processing}
	
\begin{figure}[tp]
		\centering
		\includegraphics[width=0.65\linewidth]{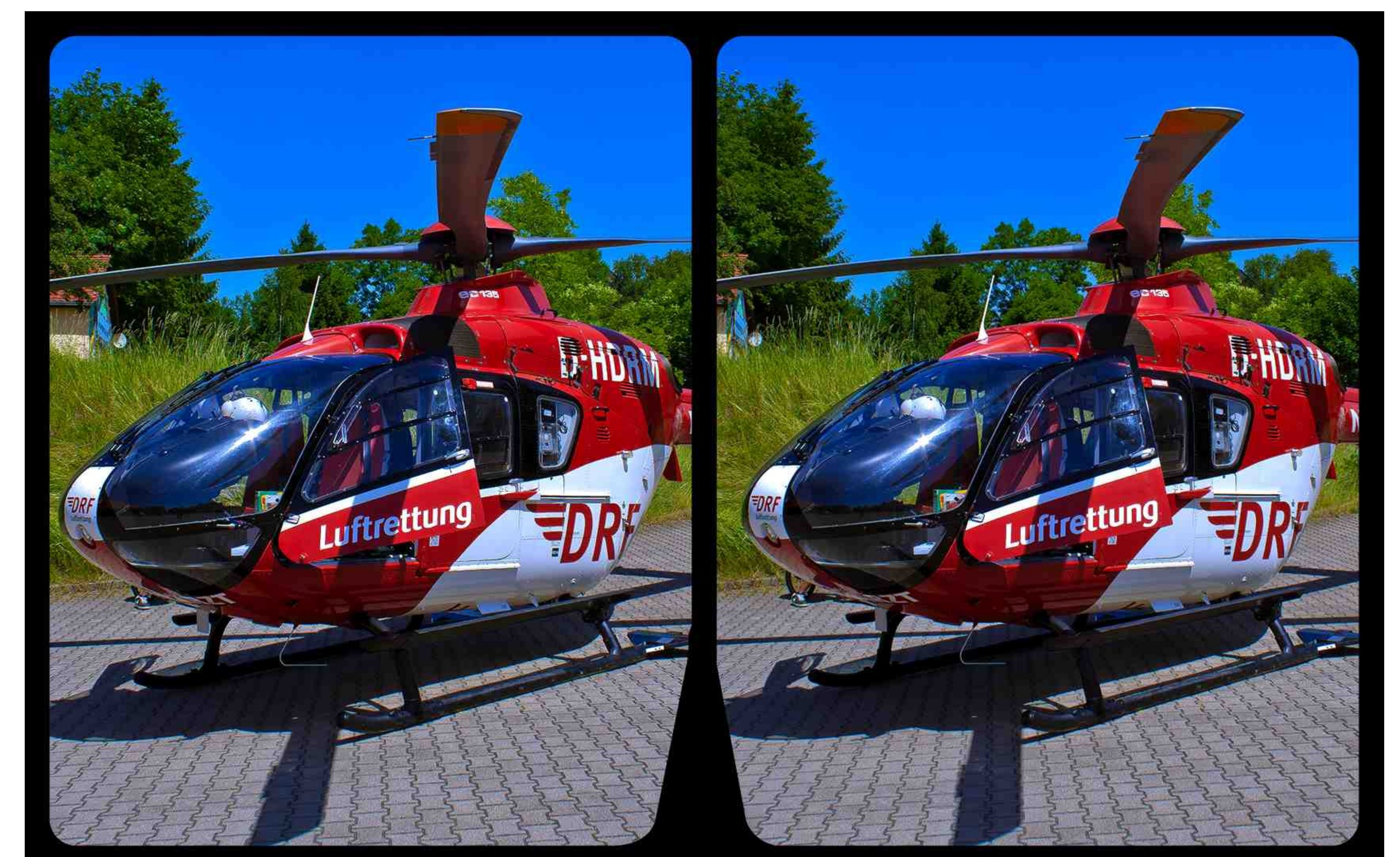}
		\caption{An example stereogram collected from Flickr.}
		\label{fig7}
	\end{figure}

	We manually collected 1024 pairs of RGB stereo images  from Flickr using tags such as stereophotography, stereoscopic and cross-eye 3D. All images are stereograms (as shown in Fig. \ref{fig7}) taken by amateur photographers using dual lens or dual cameras.
	
	First, we cut each stereogram into a pair of images and crop their black margins. The resulting left and right images from a stereogram are exchanged since the stereograms are provided in a cross-eye mode. Note that, these stereo image pairs are originally shifted to a common focus plane by the amateurs to produce a perception of 3D for viewers. In other words, both positive and negative disparities exist in these image pairs. Therefore, we roughly shift these images back to ensure that zero disparity corresponds to infinite depth. For close shots, since regions with infinite depth are unavailable, we just shift these images to make the minimum disparity larger than a threshold (empirically set to 40 pixels in our dataset). Finally, we crop each resulting image to multiple of 12 pixels on both axes following \cite{Agustsson2017NTIRE}. 
	
	\subsubsection{Comparison to Existing Datasets}
	
	The Flickr1024 dataset is compared to three widely used stereo datasets including Middlebury, KITTI 2012 and KITTI 2015. CNNIQA \cite{Kang2014Convolutional} and SR-metric \cite{Ma2017Learning} are used to evaluate the image quality. It can be observed from Table~\ref{tab4} that our Flickr1024 dataset is larger than other datasets by at least 5 times. Besides, the pixel per image (ppi) value of our Flickr1024 dataset is nearly 2 times those of the KITTI 2012 and KITTI 2015 datasets. Although the Middlebury dataset has the highest ppi value, the number of image pairs in this dataset is very limited. Further, our Flickr1024 dataset achieves comparable or even better CNNIQA and SR-metric values than other datasets, which demonstrates the good image quality of our Flickr1024 dataset.
	
	\subsection{Experimental Results}
	\label{Sec5.4}
	
	\subsubsection{Datasets}
	
	For training, we followed \cite{Jeon2018Enhancing} and downsampled 60 Middlebury \cite{Scharstein2014High} images by a factor of 2 to generate HR images. We further used Flickr1024 as the additional training data for our PASSRnet. For test, we used 5 images from the Middlebury dataset \footnote{Middlebury: cloth2, motorcycle, piano, pipes and sword2.}, 20 images from the KITTI 2012 dataset \cite{Geiger2012Are} \footnote{KITTI2012: 000000\_10 to 000019\_10.} and 20 images from the KITTI 2015 dataset \cite{Menze2015Object} \footnote{KITTI2015: 000000\_10 to 000019\_10.}. For validation, we selected another 20 images from the KITTI 2012 dataset.
	
	\begin{table}[t]
		\caption{Comparison between the Middlebury, KITTI 2012, KITTI 2015 and Flickr1024 datasets. Only the training sets of the KITTI 2012 and KITTI 2015 datasets are considered.}
		\label{tab4}
		\begin{center}
			\footnotesize
			\setlength{\tabcolsep}{0.6mm}{
				\begin{tabular}{|l|c|c|c|c|}
					\hline 
					Dataset & image pairs & ppi & CNNIQA ($\downarrow$) & SR-metric ($\uparrow$)
					\tabularnewline
					\hline
					Middlebury & 65 & \textbf{3511605} & 20.18 & 6.01 \tabularnewline
					\hline
					KITTI 2012 & 194 & 462564 & 20.32  & \textbf{7.15}  \tabularnewline
					\hline
					KITTI 2015 & 200 & 465573 & 22.86 & 7.06  \tabularnewline
					\hline
					Flickr1024  & \textbf{1024} & 800486 & \textbf{19.75}  & 7.12 \tabularnewline
					\hline
			\end{tabular}}
		\end{center}
	\end{table}
	
	\subsubsection{Implementation Details}
	
	\noindent \textbf{Training Details.}
	We first downsampled HR images using bicubic interpolation to generate LR images, and then cropped $30\times90$ patches with a stride of 20 from these LR images. Meanwhile, their corresponding patches in HR images were cropped. The horizontal patch size was set to 90 to cover most disparities ($\sim$96\%) in our training dataset. These patches were randomly flipped horizontally and vertically for data augmentation. Note that, rotation was not performed for augmentation, since it would destroy the epipolar constraints.
	
	Our PASSRnet was implemented in Pytorch on a PC with an Nvidia GTX 1080Ti GPU. All models were optimized using the Adam method \cite{Kingma2015Adam} with $\beta_{1}=0.9$, $\beta_{2}=0.999$ and a batch size of 32. The initial learning rate was set to $2\!\times\!10^{-4}$ and reduced to half after every 30 epochs. The training was stopped after 80 epochs since more epochs do not provide further improvement.
	
	\noindent \textbf{Metrics.}
	We used peak signal-to-noise ratio (PSNR) and SSIM to test SR performance. Following \cite{Jeon2018Enhancing}, we cropped borders to achieve fair comparison.
	
	\begin{table*}[ht]
		\caption{Comparative results achieved on the KITTI 2015 dataset by PASSRnet with different settings for $4\times$ SR.}
		\label{tab5}
		\begin{center}
			\footnotesize
			\setlength{\tabcolsep}{2mm}{
				\begin{tabular}{|l|c|c|c|c|c|}
					\hline 
					Model & Input & PSNR & SSIM & Params. & Time
					\tabularnewline
					\hline
					PASSRnet with single input & Left & 25.27 & 0.770  & \textbf{1.32M} & \textbf{114ms} 
					\tabularnewline
					PASSRnet with replicated inputs & Left-Left & 25.29 & 0.771 & 1.42M & 176ms 
					\tabularnewline
					\hline
					PASSRnet without residual manner & Left-Right & 25.40 & 0.774  & 1.42M&176ms
					\tabularnewline
					PASSRnet without atrous convolution  & Left-Right & 25.38  & 0.773   &1.42M & 176ms 
					\tabularnewline
					\hline
					PASSRnet without parallax-attention module & Left-Right & 25.28  & 0.771   & \textbf{1.32M} & 135ms
					\tabularnewline
					PASSRnet without transition block & Left-Right & 25.36 & 0.773 & 1.34M & 160ms 
					\tabularnewline
					\hline				
					PASSRnet  & Left-Right & \textbf{25.43} & \textbf{0.776} & 1.42M & 176ms 
					\tabularnewline
					\hline
			\end{tabular}}
		\end{center}
	\end{table*}
	
	\subsubsection{Ablation Study}
	
	\noindent \textbf{Single Input vs. Stereo Input.}
	Compared to single images, stereo image pairs provide additional information observed from a different viewpoint. To demonstrate the effectiveness of stereo information for SR, we removed the parallax-attention module from our PASSRnet and retrained the network with single images (\emph{i.e.}, the left images). For comparison, we also used pairs of replicated left images as the input to the original PASSRnet. It can be observed from Table~\ref{tab5} that the network trained with single images suffers a decrease of 0.16 dB (from 25.43 to 25.27) in terms of PSNR. If pairs of replicated left images are fed to the original PASSRnet, the PSNR value is decreased to 25.29 dB. Without extra information introduced by stereo images, our PASSRnet with replicated images achieves comparable performance to the network trained with single images. This clearly demonstrates that stereo information contributes to the performance improvement of our PASSRnet.
	
	\noindent \textbf{Residual ASPP Module.}
	Residual ASPP module is used in our network to extract multi-scale features. To demonstrate its effectiveness, two variants were introduced. First, to test the effectiveness of residual connections, we removed them to obtain a cascading ASPP module. Then, to test the effectiveness of atrous convolutions, we replaced them with ordinary convolutions. From the comparative results shown in Table~\ref{tab5}, we can see that SR performance benefits from both residual connections (25.43 vs. 25.40) and atrous convolutions (25.43 vs. 25.38). That is because, residual connections enable our residual ASPP module to extract features at more scales, resulting in more representative features. Furthermore, large receptive field of atrous convolutions facilitates our PASSRnet to employ context information in a large area. Therefore, more accurate correspondence can be obtained to improve SR performance.
	
	\noindent \textbf{Parallax-attention Module.}
	Parallax-attention module is used to exploit stereo correspondence for feature aggregation. To demonstrate its effectiveness, we introduced a variant by removing the parallax-attention module and directly stacking the features produced by the residual ASPP module. It can be observed from Table \ref{tab5} that the PSNR value is decreased from 25.43 dB to 25.28 dB if parallax-attention module is removed. Without parallax-attention module, local receptive field hinders our network to capture stereo correspondence over large disparities. \textcolor{black}{By using the parallax-attention module, our network can effectively aggregate features from left and right images with global receptive field along the epipolar line to achieve better performance.}
	
	\noindent 
	\textbf{Transition Block.}
	Transition block is used in the parallax-attention module to alleviate the training conflict in shared layers. To demonstrate the effectiveness of transition block, we removed it from our parallax-attention module and retrained the network. It can be observed from Table \ref{tab5} that the PSNR value is decreased from 25.43 dB to 25.36 dB without transition block. That is because, the transition block enhances task-specific feature learning and alleviates training conflict in shared layers.
	
	\begin{table}[t]
		\caption{Comparison between our PAM and the cost volume formation for $4\times$ SR. FLOPs are calculated on $128\!\times\!128\!\times\!64$ input features, while Time/PSNR/SSIM values are achieved on the KITTI 2015 dataset.}
		\label{tab6}
		\begin{center}
			\footnotesize
			\setlength{\tabcolsep}{1mm}{
				\begin{tabular}{|l|c|c|c|c|c|}
					\hline 
					Model & Params. & FLOPs & Time & PSNR & SSIM
					\tabularnewline
					\hline
					PAM & 94K & $1\!\times$ & $1\!\times$ & 25.43 & 0.776 \tabularnewline
					\hline
					Cost Volume  & 221K & $151\!\times$ & $\!1.5\times$ & 25.23 & 0.768 
					\tabularnewline
					\hline
			\end{tabular}}
		\end{center}
	\end{table}
	
	\noindent \textbf{PAM vs. Cost Volume}
	Cost volume and 3D convolutions are commonly used to obtain stereo correspondence \cite{Kendall2017End,Chang2018Pyramid}. To demonstrate the efficiency of our PAM in stereo correspondence generation, we replaced parallax-attention module with a 4D cost volume followed two 3D convolutional layers ($3\times3\times3$). Comparative results achieved on the KITTI 2015 dataset are listed in Table~\ref{tab6}. As compared to the cost volume formation, the number of parameters and FLOPs of our parallax-attention module are reduced by over 2 and 150 times, respectively. Using the parallax-attention module, our PASSRnet achieves better SR performance (\emph{i.e.}, PSNR value is increased from 25.23 dB to 25.43 dB) and higher efficiency (\emph{i.e.}, running time is decreased by 1.5 times). That is because, two 3D convolutional layers are insufficient to capture long-range correspondence within the cost volume. However, adding more layers will lead to a significant increase of computational cost.
	
	\begin{table}[t]
		\caption{Comparative results achieved on KITTI 2015 by our PASSRnet trained with different losses for $4\times$ SR.}
		\label{tab7}
		\begin{center}
			\footnotesize
			\setlength{\tabcolsep}{0.8mm}{
				\begin{tabular}{|c|cccc|c|c|}
					\hline 
					Model & $\mathcal{L}_{SR}$ & $\mathcal{L}_{photometric}$ & $\mathcal{L}_{smooth}$ & $\mathcal{L}_{cycle}$ & PSNR & SSIM 
					\tabularnewline
					\hline
					PASSRnet & \checkmark   &   &   &  & 25.35 & 0.771  \tabularnewline
					PASSRnet & \checkmark & \checkmark &   &  & 25.38 & 0.773 
					\tabularnewline
					PASSRnet & \checkmark & \checkmark & \checkmark &  & 25.40 & 0.774 
					\tabularnewline
					PASSRnet & \checkmark & \checkmark & \checkmark & \checkmark& \textbf{25.43}& \textbf{0.776} \tabularnewline
					\hline
			\end{tabular}}
		\end{center}
	\end{table}
	
	\noindent \textbf{Losses}
	To test the effectiveness of our losses, we retrained PASSRnet using different losses. It can be observed from Table \ref{tab7} that the PSNR value of our PASSRnet is decreased from 25.43 dB to 25.35 dB if PASSRnet is trained with only SR loss. \textcolor{black}{That is because, with only this loss, our PAM 
	cannot capture accurate stereo correspondence for feature aggregation.} If photometric loss, smoothness loss and cycle loss are added, the performance is gradually improved from 25.35/0.771 to 25.43/0.776. That is because, these losses encourage our PAM to generate reliable and consistent correspondence. 
	
	\begin{table*}[ht]
		\caption{Comparative PSNR/SSIM values achieved on the Middlebury, KITTI 2012 and KITTI 2015 datasets. Results marked with * are directly copied from the corresponding paper. Note that, only 2$\times$ SR results of StereoSR are presented on the KITTI 2012 and KITTI 2015 datasets since a 4$\times$ SR model is unavailable.}
		\label{tab8}
		\begin{center}
			\footnotesize
			\setlength{\tabcolsep}{1.5mm}{
				\begin{tabular}{|c|c|ccccc||cc|}
					\hline 
					\multirow{2}{*}{Dataset} &  \multirow{2}{*}{Scale} & \multicolumn{5}{c||}{Single Image SR} & \multicolumn{2}{c|}{Stereo Image SR}
					\tabularnewline
					&  
					&SRCNN \cite{Dong2014Learning} 
					&VDSR \cite{Kim2016Accurate}  
					&DRCN \cite{Kim2016Deeply}
					&LapSRN \cite{Lai2017Deep} 
					&DRRN  \cite{Tai2017Image} 
					&StereoSR \cite{Jeon2018Enhancing} 
					& Ours 
					\tabularnewline
					\hline
					\multirow{2}{*}{\tabincell{c}{Middlebury\\(5 images)}} & $\times$2 &32.05/0.935&32.66/0.941&32.82/0.941&32.75/0.940&32.91/0.945&33.05/0.955*&\textbf{34.05/0.960} 
					\tabularnewline
					& $\times$4 &27.46/0.843&27.89/0.853&27.93/0.856&27.98/0.861&27.93/0.855&26.80/0.850*&\textbf{28.63/0.871} 
					\tabularnewline
					\hline
					
					\multirow{2}{*}{\tabincell{c}{KITTI 2012\\(20 images)}} & $\times$2 &29.75/0.901&30.17/0.906&30.19/0.906&30.10/0.905&30.16/0.908&30.13/0.908&\textbf{30.65/0.916} 
					\tabularnewline
					& $\times$4 &25.53/0.764&25.93/0.778&25.92/0.777&25.96/0.779&25.94/0.773&-&\textbf{26.26/0.790} 
					\tabularnewline
					\hline
					\multirow{2}{*}{\tabincell{c}{KITTI 2015\\(20 images)}} & $\times$2 &28.77/0.901&28.99/0.904&29.04/0.904&28.97/0.903&29.00/0.906&29.09/0.909&\textbf{29.78/0.919} 
					\tabularnewline
					& $\times$4 &24.68/0.744&25.01/0.760&25.04/0.759&25.03/0.760&25.05/0.756&-&\textbf{25.43/0.776} 
					\tabularnewline
					\hline
			\end{tabular}}
		\end{center}
	\end{table*}
	\begin{figure*}[t]
		\centering
		\includegraphics[width=1\linewidth]{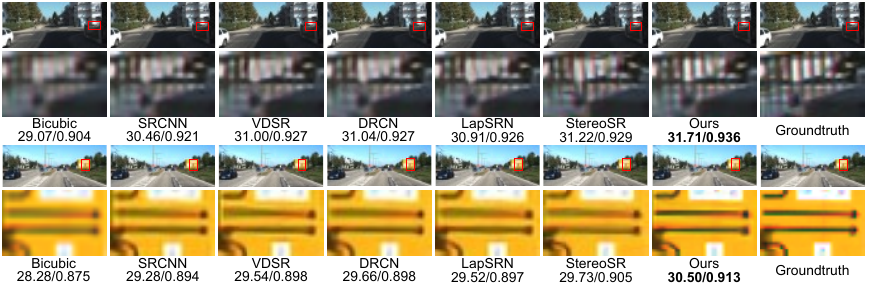}
		\caption{Visual comparison for $2\times$ SR. These results are achieved on ``test\_image\_013'' of the KITTI 2012 dataset and ``test\_image\_019'' of the KITTI 2015 dataset.}
		\label{fig9}
	\end{figure*}
	
	\subsubsection{Comparison to State-of-the-arts}
	
	We compared our PASSRnet to recent CNN-based single image SR and stereo image SR methods, including SRCNN \cite{Dong2014Learning}, VDSR \cite{Kim2016Accurate}, DRCN \cite{Kim2016Deeply}, LapSRN \cite{Lai2017Deep}, DRRN \cite{Tai2017Image} and StereoSR \cite{Jeon2018Enhancing}. The codes provided by the authors of these methods were used to conduct experiments. Note that, similar to \cite{Jeon2018Enhancing,Ahn2018Fast}, EDSR \cite{Lim2017Enhanced}, RDN \cite{Zhang2018Residual} and D-DBPN \cite{Haris2018Deep} are not included in our comparison since their model sizes are larger than our PASSRnet by at least 8 times.	
	
	\noindent \textbf{Quantitative Results.}
	The quantitative results are shown in Table~\ref{tab8}. It can be observed that our PASSRnet achieves the best performance on the Middlebury, KITTI 2012 and KITTI 2015 datasets. Specifically, our PASSRnet outperforms the best single image SR method (\emph{i.e.}, DRRN) by 1.04 dB in terms of PSNR on the Middlebury dataset for $2\times$ SR. Moreover, the PSNR value achieved by our network is higher than that of StereoSR by 1.00 dB. That is because, more accurate correspondence can be captured by our PAM. 
	
	\noindent \textbf{Qualitative Results.}
	Figure~\ref{fig9} illustrates the qualitative results achieved on \textcolor{black}{two} scenarios. It can be observed from zoom-in regions that single image SR methods cannot reliably recover missing details and suffer from obvious artifacts. In contrast, our PASSRnet uses stereo correspondence to produce finer details with fewer artifacts, such as the railings and stripe in Fig.~\ref{fig9}. Compared to StereoSR \cite{Jeon2018Enhancing}, our PASSRnet captures more accurate stereo correspondence for feature aggregation. Consequently, visually more promising results are produced.
	
	\subsubsection{Flexibility to Disparity Variations}
	We furhter tested the flexibility of our PASSRnet with respect to large disparity variations.

	\begin{table}[bt]
		\caption{Comparison between \textcolor{black}{our PASSRnet and StereoSR \cite{Jeon2018Enhancing}} on stereo images with different resolutions for 2$\times$ SR.}
		\begin{center}
			\footnotesize
			\setlength{\tabcolsep}{1.2mm}{
				\begin{tabular}{|l|cc|cc|}
					\hline 
					\multirow{2}{*}{Resolution} &  \multicolumn{2}{c|}{StereoSR \cite{Jeon2018Enhancing}} & \multicolumn{2}{c|}{Ours}
					\tabularnewline
					& PSNR & FLOPs & PSNR & FLOPs
					\tabularnewline
					\hline
					High ($500\times500$)  & 39.27 &\textcolor{black}{1}$\times$ & \textcolor{black}{\textbf{41.45}}($\uparrow2.18$) & \textcolor{black}{0.57}$\times$
					\tabularnewline
					Middle ($100\times100$) & 34.21 & \textcolor{black}{1}$\times$ & \textcolor{black}{\textbf{35.04}}($\uparrow0.83$) & \textcolor{black}{0.58}$\times$
					\tabularnewline
					Low ($20\times20$) & 29.48 & \textcolor{black}{1}$\times$ & \textcolor{black}{\textbf{29.88}}($\uparrow0.40$) & \textcolor{black}{0.36}$\times$
					\tabularnewline
					\hline
					Average            & 34.32 & \textcolor{black}{1}$\times$ & \textcolor{black}{\textbf{35.46}}($\uparrow1.14$) & \textcolor{black}{0.50}$\times$
					\tabularnewline
					\hline
			\end{tabular}}
		\end{center}
		\label{tab9}
	\end{table}
	\begin{table}[t]
		\caption{Comparison between \textcolor{black}{our PASSRnet and StereoSR \cite{Jeon2018Enhancing}} on stereo images with different baselines for 2$\times$ SR.}
		\begin{center}
			\footnotesize
			\setlength{\tabcolsep}{1.5mm}{
				\begin{tabular}{|l|cc|cc|}
					\hline 
					\multirow{2}{*}{Baseline} &  \multicolumn{2}{c|}{StereoSR \cite{Jeon2018Enhancing}} & \multicolumn{2}{c|}{Ours}
					\tabularnewline
					& PSNR & SSIM & PSNR & SSIM
					\tabularnewline
					\hline
					Large  & 37.13  & 0.9605 & \textbf{38.43}($\uparrow$1.30)  & \textbf{0.9690}($\uparrow$0.085)  
					\tabularnewline
					Medium & 37.35  & 0.9628 & \textbf{38.50}($\uparrow$1.15) & \textbf{0.9692}($\uparrow$0.064)
					\tabularnewline
					Short & 37.36  & 0.9628  & \textbf{38.50}($\uparrow$1.14) & \textbf{0.9693}($\uparrow$0.065)
					\tabularnewline
					\hline
					Average & 37.28  & 0.9620  & \textbf{38.48}($\uparrow$1.20) & \textbf{0.9692}($\uparrow$0.072)
					\tabularnewline
					\hline
			\end{tabular}}
		\end{center}
		\label{tab10}
	\end{table}	

	\begin{figure}[t]
		\centering
		\includegraphics[width=0.95\linewidth]{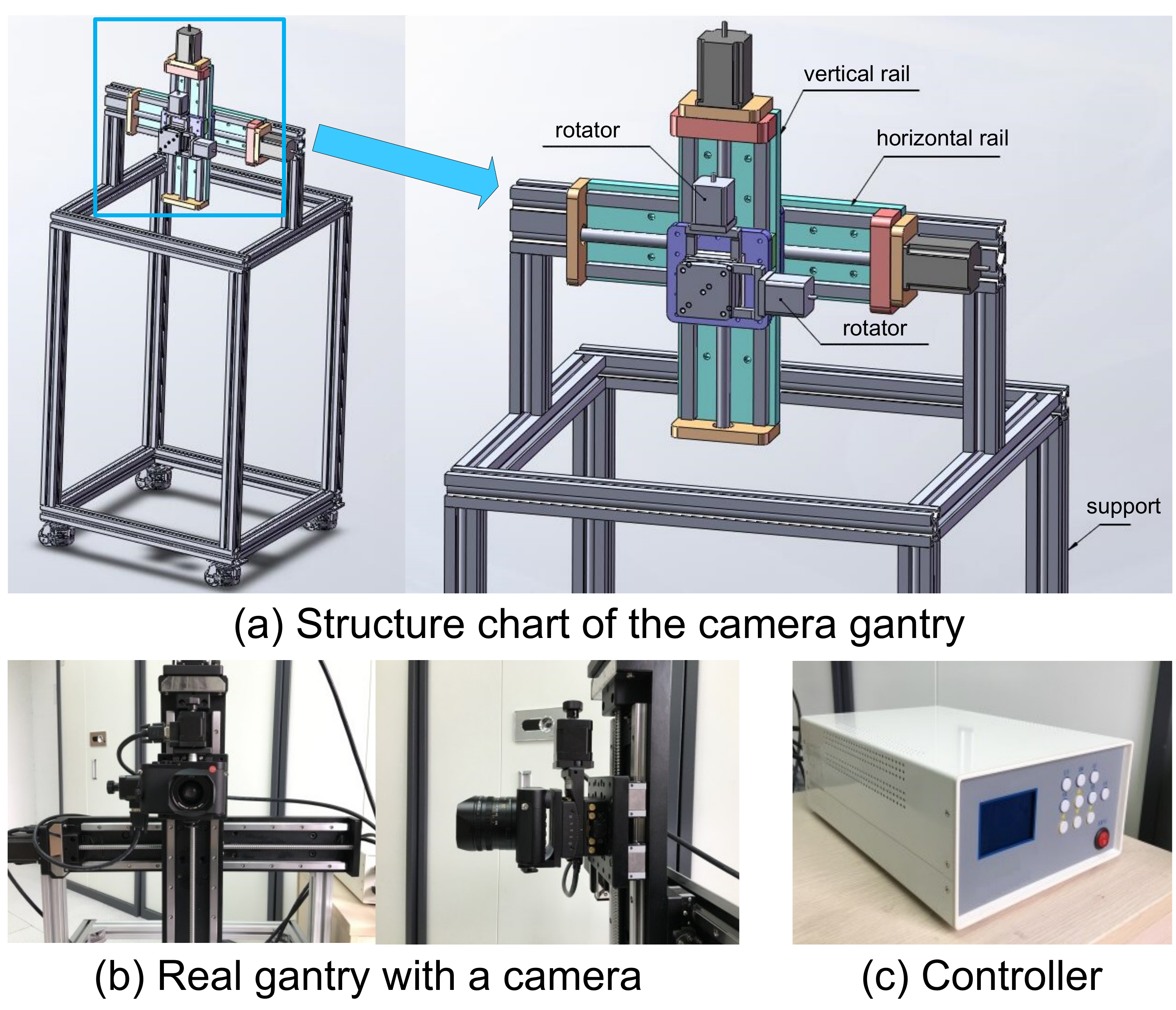}
		\caption{An illustration of the device used for stereo image acquisition.}
		\label{fig10}
	\end{figure}
	
	\noindent \textbf{Resolutions.}
	We collected 10 close-shot stereo images (with disparities larger than 200) from Flickr and resized them to different resolutions to test the flexibility of our network. Results achieved on images with different resolutions are shown in Table \ref{tab9}. 
	
	It can be observed that our PASSRnet is significantly better than StereoSR in terms of efficiency (\emph{i.e.}, FLOPs) on low resolution images. Meanwhile, our PASSRnet outperforms StereoSR by a large margin in terms of PSNR on high resolution images. That is because, StereoSR needs to perform padding for images with horizontal resolution lower than 64 pixels, which involves unnecessary calculations. For high resolution images, the fixed maximum disparity hinders StereoSR to capture longer-range correspondence. Therefore, the SR performance of StereoSR is limited.
	
	\noindent \textbf{Baselines.}
	We used a view-by-view scanning scheme as in \cite{Wang2019Selective} to obtain stereo image pairs with different baselines. The device used for image acquisition is shown in Fig. \ref{fig10}. Specifically, we first installed a camera on the gantry and acquired an image of a static scene. This image was referred to as the left image. Then, the camera was controlled to move rightward along the horizontal rail with different distances. The images acquired at different locations were referred to as right images with different baselines. Five different scenes were used for image acquisition. For each scene, we obtained three stereo image pairs with short baselines (with disparities around 20 pixels), medium baselines (with disparities around 60 pixels) and large baselines (with disparities around 180 pixels). Results achieved by StereoSR \cite{Jeon2018Enhancing} and our PASSRnet on these stereo image pairs with different baselines are shown in Table \ref{tab10}. 
	
	It can be observed that our PASSRnet outperforms StereoSR by 1.14 dB in terms of PSNR for stereo image pairs with short baselines. For stereo images pairs with large baselines, the improvement is increased to 1.30 dB. When the baseline is increased, StereoSR suffers from a more significant performance drop in terms of PSNR as compared to our PASSRnet (0.23 vs. 0.07). This clearly demonstrates that our PASSRnet is more flexible and robust to large disparity variations than StereoSR.

	\noindent \textbf{Depths.}
	We collected 30 stereo image pairs from Flickr. These image pairs cover different scenes with different depths. We divided these image pairs into three groups with small depths (with disparities around 150 pixels), medium depths (with disparities around 80 pixels) and large depths (with disparities around 20 pixels). Results achieved by StereoSR \cite{Jeon2018Enhancing} and our PASSRnet on these stereo image pairs with different depths are shown in Table~\ref{tab11}. 
	
	It can be observed that our PASSRnet achieves high improvement on stereo image pairs with small depths. That is because, the fixed maximum disparity used in StereoSR hinders longer-range correspondence to be employed. In contrast, by using PAM, our PASSRnet can get rid of setting a fixed maximum disparity and incorporate stereo correspondence under large disparity variations.

	\begin{table}[t]
		\caption{Comparison between \textcolor{black}{our PASSRnet and StereoSR \cite{Jeon2018Enhancing}} on stereo images with different depths for 2$\times$ SR.}
		\begin{center}
			\footnotesize
			\setlength{\tabcolsep}{1.2mm}{
				\begin{tabular}{|l|cc|cc|}
					\hline 
					\multirow{2}{*}{Depth} &  \multicolumn{2}{c|}{StereoSR \cite{Jeon2018Enhancing}} & \multicolumn{2}{c|}{Ours}
					\tabularnewline
					& PSNR & SSIM & PSNR & SSIM
					\tabularnewline
					\hline
					Small  & 37.60  & 0.9652 & \textbf{39.03}($\uparrow$1.43) & \textbf{0.9749}($\uparrow$0.0097)  
					\tabularnewline
					Medium & 31.08  & 0.9145 & \textbf{32.37}($\uparrow$1.29) & \textbf{0.9219}($\uparrow$0.0074)
					\tabularnewline
					Large  & 36.36  & 0.9596 & \textbf{37.55}($\uparrow$1.19) & \textbf{0.9646}($\uparrow$0.0050)
					\tabularnewline
					\hline
					Large  & 35.01  & 0.9464 & \textbf{36.32}($\uparrow$1.31) & \textbf{0.9538}($\uparrow$0.0074)
					\tabularnewline
					\hline
			\end{tabular}}
		\end{center}
		\label{tab11}
	\end{table}	
	
	\section{Conclusion}
	\label{Sec6}
	In this paper, we propose a generic parallax-attention mechanism (PAM) to capture stereo correspondence under large disparity variations. Our PAM integrates epipolar constraints with attention mechanism to calculate feature similarities along the epipolar line. Our PAM is compact and can be applied to variant stereo tasks. We apply our PAM to stereo matching and stereo image SR tasks and therefore propose two networks (namely, PASMnet and PASSRnet).
	We also introduce a new and large-scale dataset named Flickr1024 for stereo image SR. Extensive experiments have demonstrated the effectiveness of our PAM and the state-of-the-art performance of our PASMnet and PASSRnet.
	
	\ifCLASSOPTIONcompsoc
	\section*{Acknowledgments}
	\else
	\section*{Acknowledgment}
	\fi
	
	This work was partially supported by the National Natural Science Foundation of China (No. 61972435). The authors would like to thank Sascha Becher and Tom Bentz for the approval of using their cross-eye stereo photographs on Flickr.

	\bibliographystyle{unsrt}
	\bibliographystyle{IEEEtran}
	\bibliography{IEEEabrv}

	\ifCLASSOPTIONcaptionsoff
	\newpage
	\fi

	
	

	
	\begin{IEEEbiography}[{\includegraphics[width=1in,height=1.25in,clip]{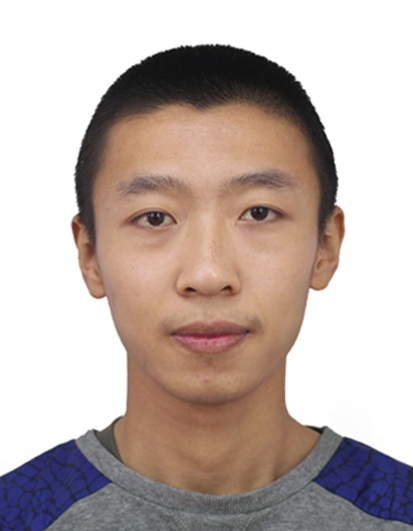}}]
		{Longguang Wang} received the B.E. degree in Electrical Engineering from Shandong University (SDU), Jinan, China, in 2015, and the M.E. degree in Information and Communication Engineering from National University of Defense Technology (NUDT), Changsha, China, in 2017. He is currently pursuing the Ph.D. degree with the College of Electronic Science and Technology, NUDT. His current research interests include low-level vision and deep learning.
	\end{IEEEbiography}
	\vspace{-5ex}
	\begin{IEEEbiography}[{\includegraphics[width=1in,height=1.25in,clip]{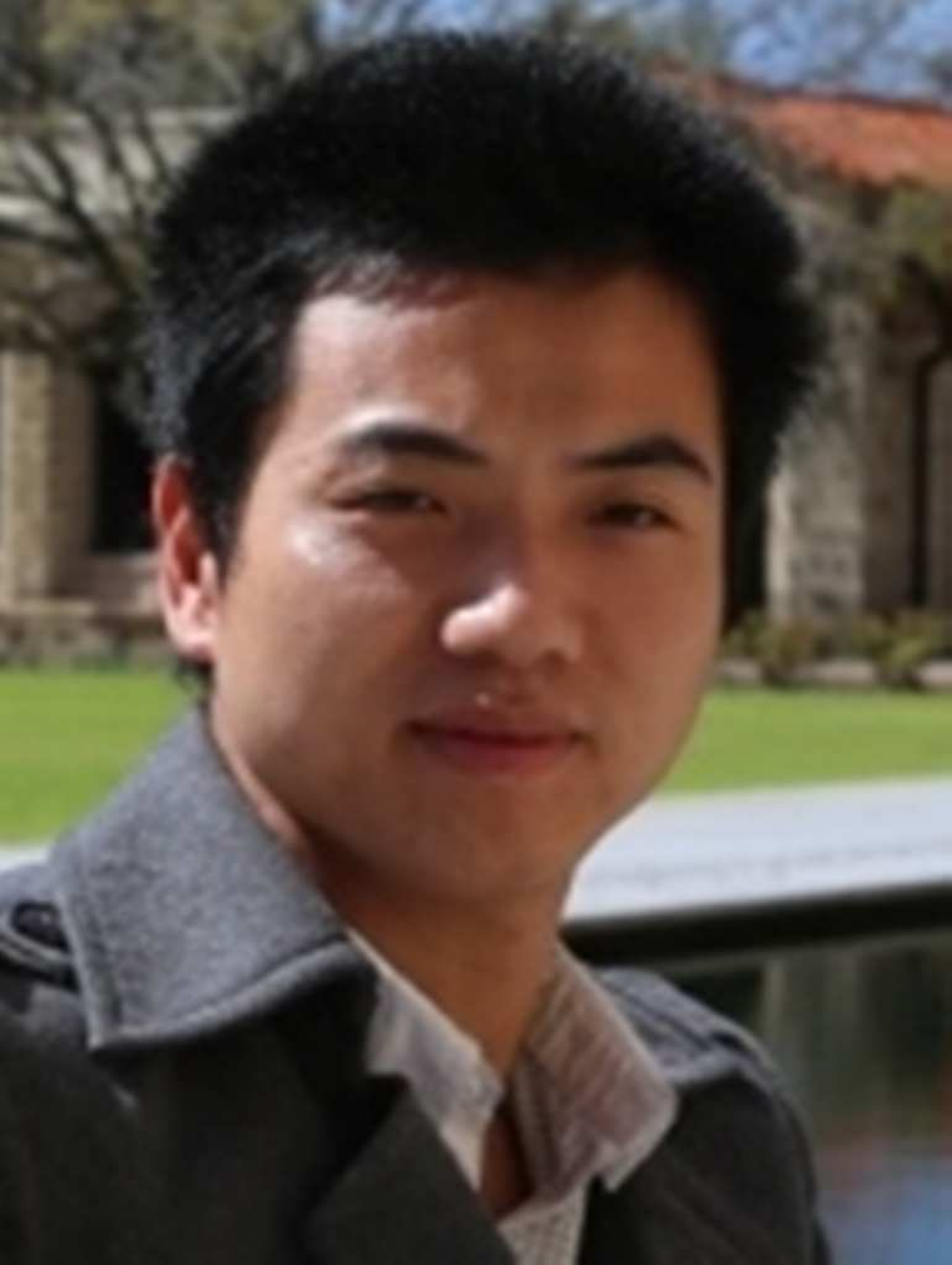}}]
	{Yulan Guo} is currently an associate professor. He received the B.Eng. and Ph.D. degrees from National University of Defense Technology (NUDT) in 2008 and 2015, respectively. He was a visiting Ph.D. student with the University of Western Australia from 2011 to 2014. He worked as a postdoctoral research fellow with the Institute of Computing Technology, Chinese Academy of Sciences from 2016 to 2018. He has authored over 90 articles in journals and conferences, such as the IEEE TPAMI and IJCV. His current research interests focus on 3D vision, particularly on 3D feature learning, 3D modeling, 3D object recognition, and scene understanding. Dr. Guo received the ACM China SIGAI Rising Star Award in 2019, Wu-Wenjun Outstanding AI Youth Award in 2019, and the CAAI Outstanding Doctoral Dissertation Award in 2016. He served as an associate editor for IET Computer Vision and IET Image Processing, a guest editor for IEEE TPAMI, and an area chair for CVPR 2021 and ICPR 2020.
	\end{IEEEbiography}
	\begin{IEEEbiography}[{\includegraphics[width=1in,height=1.25in,clip,keepaspectratio]{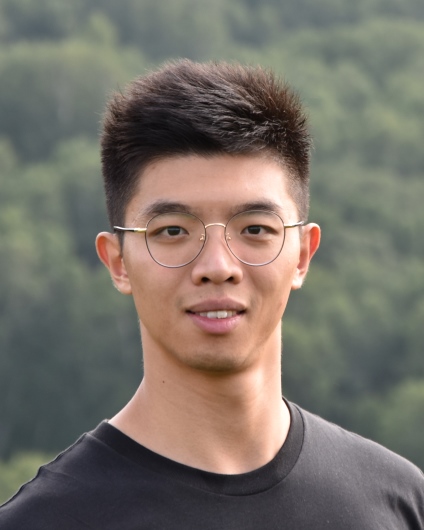}}]
		{Yingqian Wang} received the B.E. degree in Electrical Engineering from Shandong University (SDU), Jinan, China, in 2016, and the M.E. degree in Information and Communication Engineering from National University of Defense Technology (NUDT), Changsha, China, in 2018. He is currently pursuing the Ph.D. degree with the College of Electronic Science and Technology, NUDT. His current research interests focus on low-level vision, particularly on light field imaging and image super-resolution.
	\end{IEEEbiography}
	\vspace{-5ex}
	\begin{IEEEbiography}[{\includegraphics[width=1in,height=1.25in,clip,keepaspectratio]{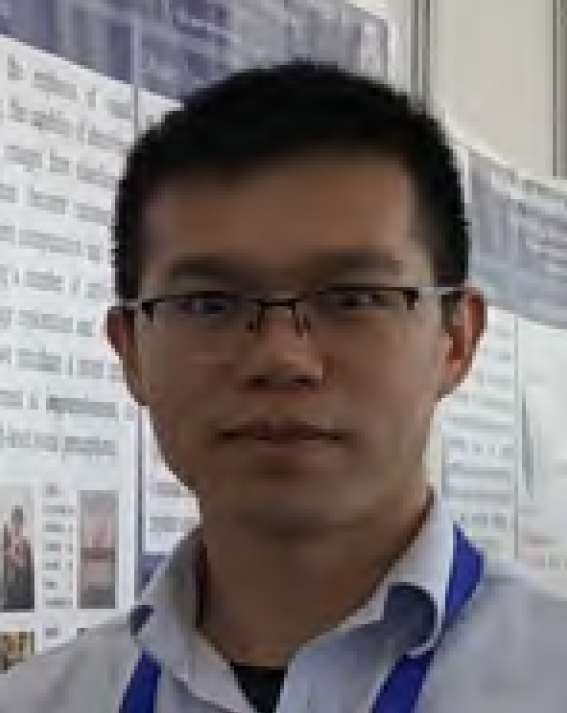}}]
		{Zhengfa Liang} received the B.Eng. degree from Tsinghua University, Beijing, China, in 2010, and the Ph.D. degree from National University of Defense Technology (NUDT), Changsha, China, in 2017. He is currently an Assistant Professor with National Key Laboratory of Science and Technology on Blind Signal Processing. His current research interests focus on 3D vision, particularly on stereo matching, depth estimation, face reconstruction, and face swapping.
	\end{IEEEbiography}
	\vspace{-5ex}
	\begin{IEEEbiography}[{\includegraphics[width=1in,height=1.25in,clip,keepaspectratio]{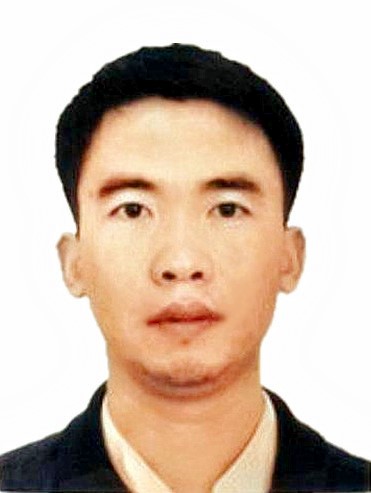}}]
		{Zaiping Lin} received the B.Eng. and Ph.D. degrees from National University of Defense Technology (NUDT) in 2007 and 2012, respectively. He is currently an Assistant Professor with the College of Electronic Science and Technology, NUDT. His current research interests include infrared image processing and signal processing.
	\end{IEEEbiography}
	\vspace{-5ex}
	\begin{IEEEbiography}[{\includegraphics[width=1in,height=1.25in,clip,keepaspectratio]{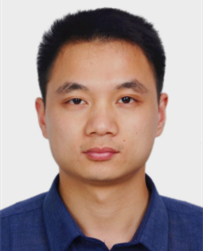}}]
		{Jungang Yang} received the B.Eng. and Ph.D. degrees  from National University of Defense Technology (NUDT),  in 2007 and 2013 respectively. He was a visiting Ph.D. student with the University of Edinburgh, Edinburgh from 2011 to 2012. He is currently an associate professor with the College of Electronic Science and Technology, NUDT. His research interests include computational imaging, image processing, compressive sensing and sparse representation. Dr. Yang received the New Scholar Award of Chinese Ministry of Education in 2012, the Youth Innovation Award and the Youth Outstanding Talent of NUDT in 2016.
	\end{IEEEbiography}
	\vspace{-5ex}
	\begin{IEEEbiography}[{\includegraphics[width=1in,height=1.25in,clip]{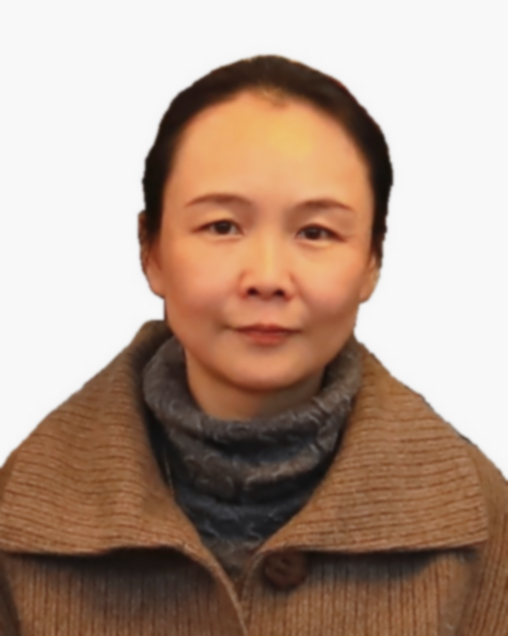}}]
		{Wei An} received the Ph.D. degree from the National University of Defense Technology (NUDT), Changsha, China, in 1999. She was a Senior Visiting Scholar with the University of Southampton, Southampton, U.K., in 2016. She is currently a Professor with the College of Electronic Science and Technology, NUDT. She has authored or co-authored over 100 journal and conference publications. Her current research interests include signal processing and image processing.
	\end{IEEEbiography}
\end{document}